\documentclass{article}

\PassOptionsToPackage{numbers}{natbib}
\usepackage[final]{neurips_2023}

\usepackage[utf8]{inputenc} 
\usepackage[T1]{fontenc}    
\usepackage{hyperref}       
\usepackage{url}            
\usepackage{booktabs}       
\usepackage{amsfonts}       
\usepackage{nicefrac}       
\usepackage{microtype}      
\usepackage{xcolor}         
\usepackage{amsmath}
\usepackage{amssymb}
\usepackage{subcaption}
\usepackage{float}
\usepackage{wrapfig}

\usepackage{graphicx}
\usepackage{listings}
\usepackage{tikz}
\usetikzlibrary{shapes,arrows}
\usetikzlibrary{patterns}
\usetikzlibrary{decorations.pathreplacing, decorations.pathmorphing, calligraphy}

\tikzstyle{rectblock} = [rectangle, draw, thick, align=center]
\tikzstyle{block} = [rectblock, rounded corners]
\tikzstyle{dashblock} = [rectangle, draw, thick, align=center, dashed]
\tikzstyle{netnode} = [circle, draw, very thick, inner sep=0pt, minimum size=0.5cm, align=center]
\tikzstyle{relunode} = [rectangle, draw, very thick, inner sep=0pt, minimum size=0.5cm]
\tikzstyle{line} = [draw, very thick, -latex']
\tikzstyle{arrow} = [draw, ->, very thick]
\tikzstyle{attention} = [arrow, bend right]
\tikzstyle{mapsto} = [draw, |->, thick]
\tikzstyle{annobrace} = [draw, thick, decorate, decoration={brace, mirror, amplitude=0.5em}]
\tikzstyle{annotext} = [pos=0.5, text width=1.5cm, anchor=north, yshift=-0.25em, align=center]

\definecolor{bpurp}{HTML}{984ea3}
\definecolor{bblue}{HTML}{377eb8}
\definecolor{bgreen}{HTML}{4daf4a}
\definecolor{borange}{HTML}{ff7f00}
\definecolor{bred}{HTML}{a50f15}

\newcommand{\doit}[1]{do\!\left(#1\right)}

\title{Passive learning of active causal strategies in agents and language models}

\author{%
  Andrew K. Lampinen \\
  Google DeepMind\\
  London, UK\\
  \texttt{lampinen@deepmind.com} \\
  \And
  Stephanie C. Y. Chan\\
  Google DeepMind\\
  London, UK\\
  \texttt{scychan@deepmind.com} \\
  \And
  Ishita Dasgupta\\
  Google DeepMind\\
  London, UK\\
  \texttt{idg@deepmind.com} \\
  \AND
  Andrew J. Nam\\
  Stanford University\\
  Stanford, CA\\
  \texttt{ajhnam@stanford.edu} \\
  \And
  Jane X. Wang\\
  Google DeepMind\\
  London, UK\\
  \texttt{wangjane@deepmind.com} \\
}

\begin{document}

\maketitle

\begin{abstract}
What can be learned about causality and experimentation from passive data? This question is salient given recent successes of passively-trained language models in interactive domains such as tool use. Passive learning is inherently limited. However, we show that purely passive learning can in fact allow an agent to learn generalizable strategies for determining and using causal structures, as long as the agent can intervene at test time. We formally illustrate that, under certain assumptions, learning a strategy of first experimenting, then seeking goals, can allow generalization from passive learning in principle. We then show empirically that agents trained via imitation on expert data can indeed generalize at test time to infer and use causal links which are never present in the training data; these agents can also generalize experimentation strategies to novel variable sets never observed in training.
We then show that strategies for causal intervention and exploitation can be generalized from passive data even in a more complex environment with high-dimensional observations, with the support of natural language explanations. Explanations can even allow passive learners to generalize out-of-distribution from otherwise perfectly-confounded training data. Finally, we show that language models, trained only on passive next-word prediction, can generalize causal intervention strategies from a few-shot prompt containing examples of experimentation, together with explanations and reasoning. These results highlight the surprising power of passive learning of active causal strategies, and may help to understand the behaviors and capabilities of language models.
\end{abstract}

\section{Introduction}

Learning from passive observational data only allows learning correlational, not causal, structure. This observation is sometimes cited as a fundamental limitation of current machine learning research \citep{pearl2018theoretical, pearl2019seven, lake2017building}. However, reinforcement learning (RL) agents can intervene on their environment, and are therefore not entirely limited. Indeed, various works have shown that RL agents can (meta-)learn to intervene on the environment to discover and exploit its causal structure \citep{nair2019causal,dasgupta2019causal,lampinen2022tell,ding2022generalizing,jiang2023learning}.

However, these prior works leave open the possibility that an agent could \emph{passively} learn a generalizable strategy for discovering and exploiting causal structure. While it is certainly necessary for an agent to intervene on the world \emph{at test time} to discover causal structure, it may be possible for the agent to learn such a strategy \emph{from purely passive, offline data}. Metaphorically, we ask ``could an agent learn to be a scientist by only reading a textbook detailing previous experiments and discoveries, and then successfully perform new experiments to discover and exploit new causal structures?''

The answer to this question is interesting for several reasons. From the perspective of causal learning, it is useful to investigate what limitations passive learning actually induces with respect to the knowledge an agentic system can later gain of the world's causal structure. Moreover, the answer is particularly salient at present because language models (LMs) --- even when trained on purely passive imitation --- show emergent capabilities such as adapting or reasoning in-context \citep{wei2022chain,nye2021show,zhou2022teaching}, or using tools to interact with the world \citep{schick2023toolformer}. These abilities are presumably driven by distributional properties of the passive data on which the LMs are trained \citep[cf.][]{chan2022data,prystawski2023think}. While LMs are often fine-tuned in active settings, e.g. using RL from Human Feedback \citep{christiano2017deep}, it is unclear whether this step is necessary to unlock causal abilities; the vast majority of learning occurs during the passive pretraining phase \citep[e.g.][]{zhou2023lima}. Thus, understanding whether passive learning can lead to generalizable causal strategies can contribute to understanding the capabilities and limitations of language models. 

In this work, we make the following contributions:
\begin{itemize} \setlength\itemsep{0.05em}
  \vspace{-0.33em}
  \item We formally show that it is possible for an agent to learn a generalizable strategy for exploiting certain causal structures from passive data, as long it can intervene at test time.
  \item We demonstrate empirically that an agent trained via imitation (behavioral cloning) on expert data from a simple causal intervention task indeed can generalize at test time to infer and use causal links that are never present in the training data.
  \item We show that passively-trained agents can also generalize strategies for adaptive experimentation to experiment optimally in a novel situation.
  \item We show that strategies for causal intervention and exploitation can be learned from passive data even in a more complex environment with pixel observations and relational structure.
  \item We further show that natural language explanations support learning and generalizing causal strategies from passive data, and can enable generalization out-of-distribution, even from otherwise totally confounded training data.
  \item Finally, we show that language models, trained only on passive next-word prediction, can generalize causal intervention strategies from a few-shot prompt containing explanations.
\end{itemize}

\section{Related work}

Here, we attempt to briefly situate our contributions within the extensive prior literature on causal structure induction, language models, and online and offline RL agents.

A variety of works have considered the problem of causal graph structure induction from data \citep{glymour2019review,nishikawa2022bayesian,peters2012identifiability}. While in a few cases the causal graph is identifiable from observational data alone \citep{peters2014identifiability}, observations are generally not sufficient \citep[e.g.][]{xia2021causal}, so most works also incorporate at least some interventional data. However, there is some flexibility in the interventional information required; for example,  \citet{faria2022differentiable} show that a deep learning system can infer causal relations from data in which interventions are present, but latent (not directly observed). Perhaps most relevantly, \citet{ke2022learning} show that a transformer-based model can learn to induce causal structures from observational and interventional data. Our work moves beyond theirs in demonstrating that an agent can passively learn both to actively collect its own interventional data, and to use implicit inferences about causal structure to achieve goals --- and by connecting these observations to recent progress in LMs.

Many perspectives on the limitations of LMs correspond to causal arguments. For example, the ``octopus test'' of \citet{bender2020climbing} is an argument about causal structure in the world that is latent in language. Other works have expressed skepticism about the possibility of LMs (or other models) learning causal strategies from passive data \citep[e.g.][]{ortega2021shaking,pearl2018theoretical}. Even works that take more optimistic perspectives tend to focus on memorization of observed causal structure --- e.g. \citet{kiciman2023causal} study LM knowledge about real-world causal structure, and note their potential utility as causal priors. \citet{willig2023causal} make some overlapping observations, but argue that these are simply examples of discovering correlations between causal structures in the real world and their descriptions in language, and that therefore LLMs ``simply recite the causal knowledge embedded in the data'' and thus ``we cannot expect any sort of generalization in terms of causal prowess.'' However, some works have suggested that LMs can learn implicit world models from the training data \citep{li2021implicit,li2022emergent}, suggesting that there is potential for more than purely superficial memorization. Correspondingly, we will show that it is possible to learn generalizable causal strategies from passive data, especially when given explanations that link observations to causal structure (which is particularly relevant for LMs).

Work in offline RL has also studied the ability of agents to generalize from passive data \citep{pmlr-v97-fujimoto19a,kumar2019stabilizing,agarwal2020optimistic}, but has generally used more complex offline algorithms rather than pure imitation. Some work on pure imitation has considered the causal challenges for learning an optimal policy \citep{ross2011reduction,dehaan2019causal,zhang2020causal}, but generally under a fixed causal structure. Other work has explored the generalization capabilities of agents trained from pure imitation when conditioned on information about the trajectories (such as returns) \citep{schmidhuber2019reinforcement,srivastava2019training,arulkumaran2022all,chen2021decision}. 
\citet{laskin2022context} show that passive offline imitation can learn a meta-RL algorithm that can improve an RL policy at test time.
However, none of these works focused on tasks of causal experimentation and exploitation.

Instead, work on learning causal strategies has generally relied on online RL agents. \citet{dasgupta2019causal} showed that causal reasoning can emerge implicitly from meta-RL; other approaches learn to infer causal structures explicitly as part of an meta-RL process \citep{nair2019causal,jiang2023learning}. \citet{lampinen2022tell} demonstrate that explanations can support meta-learning of causal strategies. Other works have proposed benchmarks for meta-learning causal experimentation \citep{wang2021alchemy}. However, all of these works relied on active RL. \citet{kosoy2022towards} recently considered learning of causal overhypotheses, and in addition to online RL agents evaluated behavioral cloning, but found it performed poorly; a decision transformer performed somewhat better, but only if forced to explore.
Thus, past works leave open the question of whether agents can generalize causal strategies from passive data.

\section{Passive learning can suffice}

The key idea is that an agent that can intervene at test time could correctly infer and use the causal structure, as long as it has acquired an appropriate strategy during training. One way to enable this is to teach the agent this strategy in two stages: first, explore a system using an \textit{experimentation strategy} that is guaranteed to identify the causal structure; second, learn an \textit{exploitation strategy} to use this identified causal structure to achieve a goal. If the agent is successfully able to execute the experimentation strategy in the first part, it will necessarily have acquired the information required. Generalizing the \textit{exploitation strategy} is directly amenable to passive learning -- given that all the relevant information has been uncovered, there is no in-principle limitation on its success.

More formally, suppose an agent \(A\) receives observations \(o_t\) at time \(t\), and is capable of taking actions \(a_t\) that intervene on a system. Assume that the system in question consists of a finite set of \(n\) random variables \(\{X_0, ..., X_n\}\) that are connected to form a causal Directed Acyclic Graph (DAG; see e.g.\ \citet{glymour2016causal}, Ch. 2). We will assume for simplicity that the variables are real-valued and that edges in the graph correspond to linear effects. We assume that independent variables (i.e., those without ancestors) are sampled from a normal distribution with mean 0 and a random variance, followed by a nonlinearity; nodes with ancestors are sampled from a normal distribution with mean corresponding to the linear effects of its inputs and again a random variance, and final nonlinearity. It is easy to adapt our argument in order to change many of these assumptions.\footnote{E.g., to accommodate interactions interevene on subsets up to the interaction order.} The agent's observations correspond to current values of the variables and (at some time steps) a goal \( g \in \mathcal{G}\), that is, \(o_t \in \mathbb{R}^n \times \mathcal{G}\). 
The agent's actions correspond to interventions that set the value of a variable \(a_t = do\left(X_i=x\right)\) for some \(i \in \{1, ..., n\}\) and \(x \in \mathbb{R}\).

Now suppose we have a dataset of expert demonstrations on varied causal structures, in which the expert first intervenes on every variable to determine the causal structure; then uses that structure to achieve some goal. For example, the trajectories of the expert might have the following form:
\[
  {\color{bred} \underbrace{a_0 = \doit{X_0\!=\!x_0}, ..., a_n =\doit{X_n\!=\!x_n}}_{\text{experimentation: determine causal structure}}}\,
  {\color{bblue} \underbrace{a_{n+1} = \doit{X_i\!=\!x^g_{n+1}}, ..., a_{n+k}= \doit{X_j\!=\!x^g_{n+k}}}_{\text{exploitation: take optimal actions to achieve goal }g\text{ given causal structure}}}
\]
In this example, the first phase --- the \textit{intervention strategy} --- is easy to learn, since it consists of a fixed sequence of actions.\footnote{N.B. we discuss this simple strategy because it is clear; it is not the most efficient \citep{eberhardt2012number}.} Thus, as long as the agent successfully learns the training data, it should reproduce this sequence on new causal structures. Once the agent has completed these interventions, it will have acquired all the information necessary to correctly infer the causal structure. With this information, the agent can in principle behave optimally in the goal-seeking remainder of the episode, i.e. implementing the \textit{exploitation strategy}. Thus, the agent is not subject to any in-principle limitations. The two core tenets are whether a) the agent is able to follow the learned \textit{exploration strategy} in new unobserved causal structures at test time, and b) the agent is able to generalize the \textit{exploitation strategy} to account for the novel outcomes of those interventions.
Below, we show empirically that an agent can indeed exhibit this kind of generalization from passive data.

\section{Methods}

\textbf{Agent:} We used a standard agent architecture: an input encoder that processes the observations, a memory that tracks and integrates information over time, and a policy that predicts actions. In analogy to the language model setting, we used a TransformerXL \citep{dai2019transformer} for the agent memory, with a sufficiently long memory to cover the expected episode length. For the encoder in the causal DAG environments we concatenated the input observations and projected through a single linear layer. For the odd-one-out environments we used a CNN for visual observation encoding and an LSTM for language (following prior work on these environments \citep{lampinen2022tell}). For the agent policy, we used a single linear layer which receives as input the state output from the Transformer memory, and produces logits over the set of possible actions. For explanation prediction output (where applicable) we used a single-layer LSTM which received the same input as the policy.

\textbf{Expert policies \& datasets:} We trained the agent on a dataset of trajectories generated by an expert policy, which behaves (near-)optimally. The specific implementation of the expert policy depends on the task, see below and Appendix \ref{appx:envs}. We place our agent in the infinite-data regime (a common assumption in language model training \citep[e.g.,][]{hoffmann2022training}), that is, the number of episodes in the dataset is larger than the number of episodes the agent trains on, so the agent does not repeat any episode.

\textbf{Language model:} For experiments with language models, we used the 70 billion parameter Chinchilla LM \citep{hoffmann2022training}; a decoder-only transformer that is trained purely on language modelling, without any downstream fine-tuning. When evaluating the LM's choice among multiple answers, we calculated the log-likelihood of each as a continuation of the current prompt, and selected the one with the highest likelihood. When generating from the model, to produce explanations and reasoning traces at test time, we used nucleus sampling \citep{holtzmancurious} with \(p = 0.8\) and \(T = 1\).

\section{Experiments} \label{sec:experiments}

\subsection{A simple causal DAG environment} \label{sec:experiments:dag}
\begin{figure}[t]
\begin{subfigure}[b]{0.25\textwidth}
\centering
\resizebox{0.98\textwidth}{!}{
\begin{tikzpicture}
\node[align=left,text width=2cm] at (-2.5, 1.5) {\bf Train data:};
\node[netnode] at (0, 1) (A) {A};
\node[netnode] at (-0.95, 0.3) (B) {B};
\node[netnode] at (-0.59, -0.81) (C) {C};
\node[netnode] at (0.59, -0.81) (D) {D};
\node[netnode] at (0.95, 0.3) (E) {E};

\draw[arrow] (D) to (A);
\draw[arrow] (C) to (D);
\draw[arrow] (C) to (E);
\draw[arrow] (D) to (B);
\draw[arrow] (E) to (A);

\draw[arrow,color=bred,dashed] (B) to (E);
\draw[very thick,color=bred] (-0.1, 0.2) to (0.1, 0.4);
\draw[very thick,color=bred] (0.1, 0.2) to (-0.1, 0.4);

\node[color=bred, text width=2.7cm] at (-2, 0.75) (train_test) {\tiny This edge would make D an\\[-0.1em]ancestor of E; not allowed\\[-0.75em]in training data.};
\draw[arrow,color=bred,bend left,dotted] ([xshift=-9]train_test.east) to (-0.15, 0.45);

\begin{scope}[shift={(0,-3)}]

\node[align=left,text width=2cm] at (-2.5, 1.5) {\bf Evaluation:};
\node[netnode] at (0, 1) (A) {A};
\node[netnode,color=bblue] at (-0.95, 0.3) (B) {B};
\node[netnode] at (-0.59, -0.81) (C) {C};
\node[netnode,color=bblue] at (0.59, -0.81) (D) {D};
\node[netnode,color=borange!80!white] at (0.95, 0.3) (E) {E};

\draw[arrow] (C) to (B);
\draw[arrow] (B) to (A);
\draw[arrow] (E) to (A);

\draw[arrow,color=bblue] (D) to (E);
\draw[arrow,color=bblue] (B) to (E);
\draw[arrow,color=bblue] (D) to (B);
\node[color=borange!80!white, text width=1.7cm] at (1.6, 0.9) {\tiny Eval goal:\\[-0.5em]maximize E.};

\node[color=bblue, text width=2.25cm] at (-2, -1) (d_ancest) {\tiny At eval time, D is\\[-0.1em]the most impactful\\[-0.75em]ancestor of E.};
\draw[arrow,color=bblue,bend right,dotted] ([xshift=-18,yshift=7]d_ancest.south east) to ([xshift=-1,yshift=-1]D.south west);
\end{scope}
\end{tikzpicture}}
\caption{Train vs.\ eval.\ DAGs}\label{fig:dag:task}
\end{subfigure}%
\begin{subfigure}[b]{0.3\textwidth}
\centering
\includegraphics[width=\textwidth]{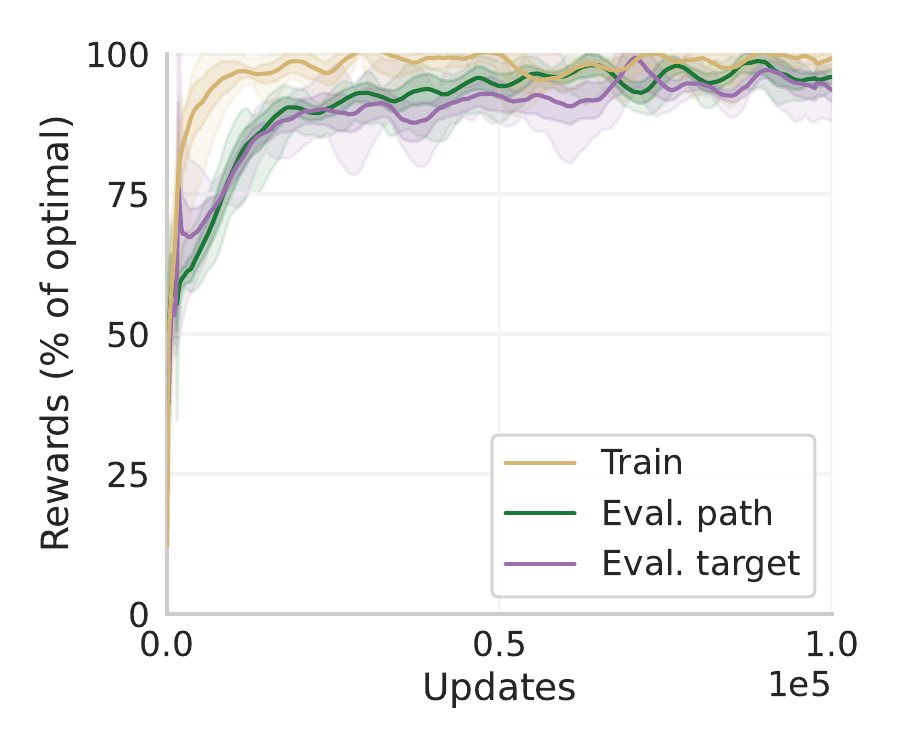}
\vskip-0.25em 
\caption{Rewards}\label{fig:dag:rewards}
\end{subfigure}
\begin{subfigure}[b]{0.445\textwidth}
\centering
\includegraphics[width=\textwidth]{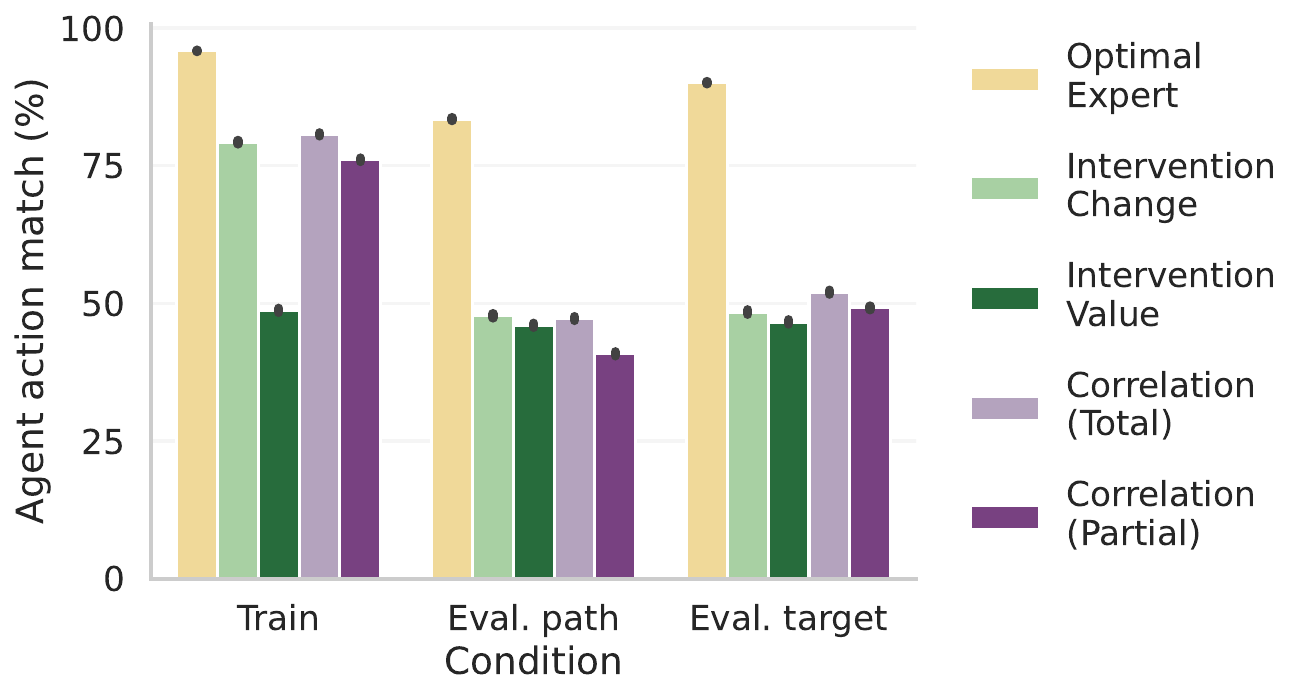}
\caption{Action analysis}\label{fig:dag:actions}
\end{subfigure}
\caption{The causal DAG environment and experimental results. (\subref{fig:dag:task}) The constraints imposed on the causal DAG structures when creating the training dataset, and when evaluating the agent on test structures. During training, D is not allowed to be an ancestor of E (even indirectly), though they may be correlated due to confounding variables. In evaluation environments, D is the most impactful ancestor of E (see text). (\subref{fig:dag:rewards}) Rewards obtained by the agent when evaluated in interactive settings, as a percentage of the optimal rewards. In both evaluation settings, the agent still achieves close to optimal reward. (\subref{fig:dag:actions}) Analyzing the agent's behavior in more detail, by plotting the proportion of the agent's actions that match the optimal behavior, or baselines based on heuristics from the interventions or correlational statistics. The agent matches the optimal strategy significantly more closely than it matches the heuristic baselines. (Error bars are 95\%-CIs; 5 seeds per condition.)} 
\label{fig:dag}
\vskip-1em
\end{figure}
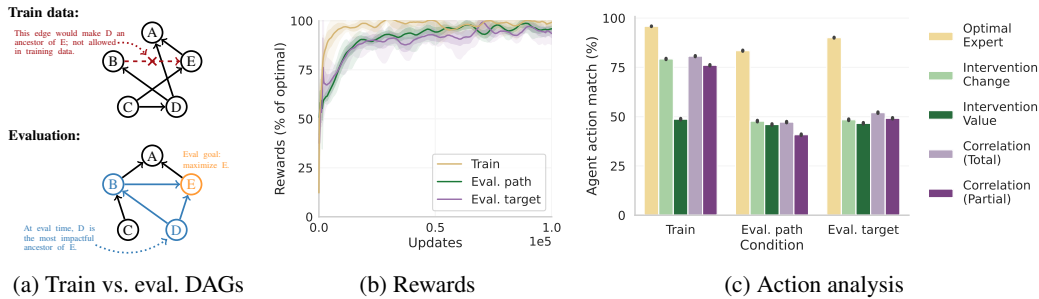

Our first experiments take place in a simple causal DAG setting, inspired by the argument above and prior work on RL agents learning causal strategies \citep{dasgupta2019causal}. The environment consists of a causal DAG over \(n = 5\) variables (A, B, C, D, E), with the causal graph, edge weights, and noise variance for each node resampled on each episode. Each episode consists of several segments: first, the agent is given \(n\) steps to explore the graph --- enough to intervene on each variable and observe the consequences --- and a second phase in which the agent is cued to try to maximize a particular variable. During this second phase, the agent is rewarded according to the value of the cued variable after its actions. In both phases, the agent observes its previous intervention and the value of all variables before and after. In the second phase, it also observes the goal cue. For details see Appx. \ref{appx:envs:dag}. 

The training dataset is generated by an optimal expert policy that first explores (by choosing a random node to start with, then iterating through subsequent nodes in order, looping back to the first node after the last, and intervening on each to determine its effect). Then, the expert plans using the inferred causal structure to identify the intervention that will optimally achieve the given goal, and takes that action. The learning agent is trained to imitate this expert dataset via behavioral cloning --- that is, by passively learning to maximize the probability of the expert actions.

The training data are generated from a subset of the possible causal DAGs, holding out some causal dependency links entirely for testing. We train the agent on expert data from DAGs in which node E is never descended (even indirectly) of node D. We then test the agent (interactively) on DAGs in which node E \emph{is} descended from node D, with sufficiently large edge weights along paths from D to E (and sufficiently small from other ancestors) that node E is most strongly influenced via node D. We evaluate the agent's ability to determine this dependency through its experiments, and then, when cued to maximize node E, to optimally intervene. Specifically, we consider two test conditions: ``eval. target,'' where node D is the optimal target to intervene on, and ``eval. path,'' where node D is somewhere on the path from the optimal target to node E (but an ancestor of node D might be a better intervention). The ``eval. target'' condition is perhaps the most difficult, as in the training data the agent never sees any structure in which intervening on node D affects node E at all. Note that this generalization split is challenging because neither the environment observations nor the agent architecture encode permutation-invariance of the variables; their values are presented in a fixed order in distinct observation entries, and the agent uses distinct actions for intervening on each variable.

Thus, the question is whether the agent can learn to infer this novel causal dependency path from its experiments, and to use it to achieve the goal --- i.e. can an agent generalize its exploitation strategy to novel causal structures? (Generalization of exploration strategy is investigated below in Sec \ref{sec:experiments:dag:adaptive}). Indeed, in Fig. \ref{fig:dag:rewards} we show that the agent rapidly learns to achieve high reward under interactive evaluation in all conditions (both training and evaluation structures). 

However, simpler heuristic strategies could achieve some rewards. We therefore compare the agent's actions \emph{during the exploitation phase} to the optimal strategy, and four baselines. The change and value baseline strategies use simpler heuristics, rather than relying on the full causal structure to achieve the goal of maximizing a given node. The value baseline simply repeats the action that during the exploration phase resulted in the highest value for the goal node. The change baseline is slightly more sophisticated; it chooses the intervention that changed the value of the goal node the most, and then flips its sign if the change was negative. We also consider two strategies that observe all the node values that the agents do, but consider only correlational statistics --- either the total correlation between two nodes, or the correlation when controlling for all other nodes. Note that these baselines and the optimal expert policy are not mutually exclusive; these heuristics are effective \emph{because} they often are match the expert. However, in Fig. \ref{fig:dag:actions} we show that the agent's actions match the expert policy significantly better than any baseline, across both training and evaluation. These observations suggest that the agent is learning to use more complex causal strategies, rather than simpler heuristics or correlational statistics.

In the supplement we perform a number of follow-up experiments, showing that \emph{at test time}, it is necessary to intervene rather than just observe (Appx. \ref{app:supp_exp:no_explore_ablation}); how causal strategy generalization depends on the variety of causal DAG structures in training (Appx. \ref{app:supp_exp:more_holdouts}); and that agents can generalize causal strategies to DAGs with different functional structures, such as linear DAGs in training and nonlinear in test (Appx. \ref{app:supp_exp:dag_function_gen}).

\subsubsection{Adaptive experimentation strategies} \label{sec:experiments:dag:adaptive}

It is limiting to assume that the optimal policy requires intervening on every causal factor; that does not scale to realistic scenarios. Scientists are guided by domain knowledge that suggests which variables may be relevant to a given phenomenon. 
\begin{wrapfigure}{r}{0.4\textwidth}
\vspace{-0.4em}
\includegraphics[width=0.35\textwidth]{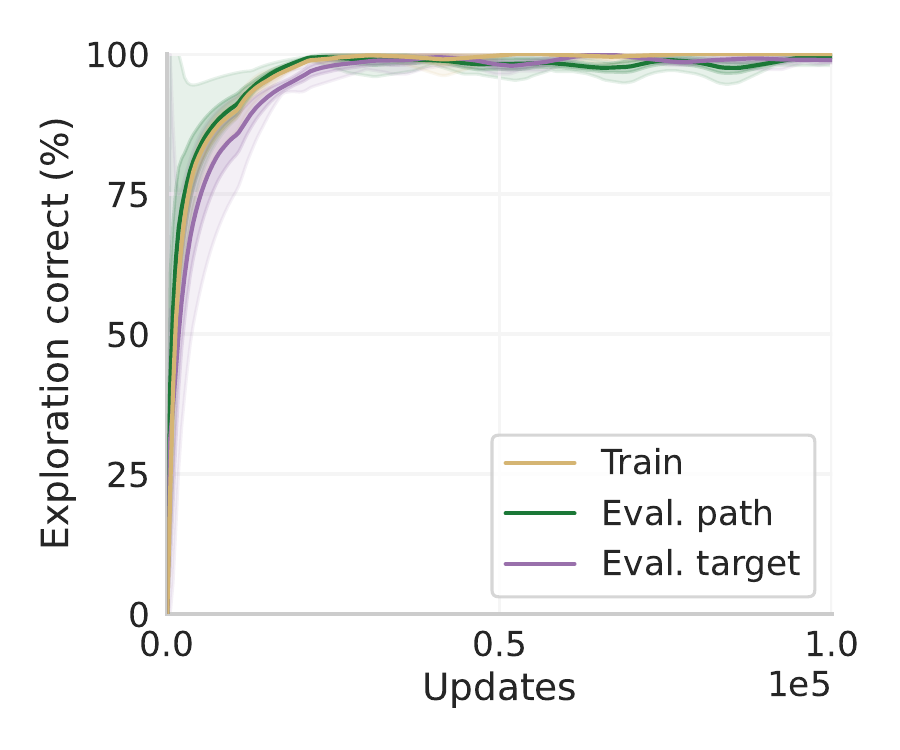}
\vspace{-0.66em}
\caption{When the agent is passively trained on expert actions that explore (intervene on) only a subset of the variables, it can generalize to explore correctly on held-out variable subsets at test time.} 
\label{fig:adaptive_dag}
\vspace{-1.2em}
\end{wrapfigure}
We therefore explored whether the passive agent can learn to adapt its experimentation procedure based on cues about which variables might matter.

To do so, we increased the variables to \(n = 10\), but restricted so that only 5 are relevant in any given episode. The other 5 are independent from everything, and the goal node to maximize is always one of the 5 relevant variables. The agent is given a many-hot cue that marks which variables are relevant currently, and the expert policy which the agent learns to imitate only intervenes on the relevant variables. In the training data, we hold out some subsets of variables such that the agent never sees experiments on these subsets, as well as holding out certain ancestor structures as above. We test the agent in settings where it encounters both a held-out set of variables, and new relations among them. 

In Fig. \ref{fig:adaptive_dag}, we plot over training the proportion of evaluation episodes in which the agent correctly uses an optimal exploration strategy; that is, in which it intervenes on each relevant variable once, without intervening on any irrelevant variable. The agent rapidly learns to generalize the experimentation strategy from the data to correctly experiment when cued with sets of variables never encountered in training. It then goes on to successfully use the results of these experiments to achieve the goal, and again aligns with the optimal causal policy more closely than the heuristic baselines (Appx. \ref{app:supp_exp:adaptive}).

\subsection{The odd-one-out environments}

\begin{figure}[h!]
\begin{subfigure}[b]{0.33\textwidth}
\centering
\resizebox{\textwidth}{!}{
\begin{tikzpicture}
\node[align=left,text width=8cm] at (-1, 3) {\Huge \bf Experiment trials \(\times 3\)};
\node at (-3, 0) (pretrans) {\includegraphics[width=4cm]{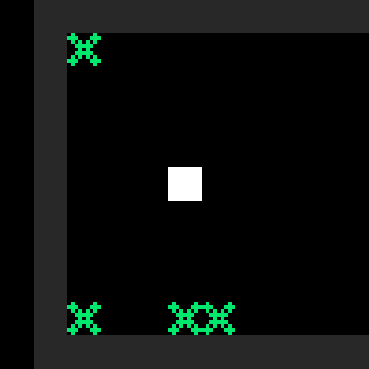}};
\node at (5, 0) (posttrans) {\includegraphics[width=4cm]{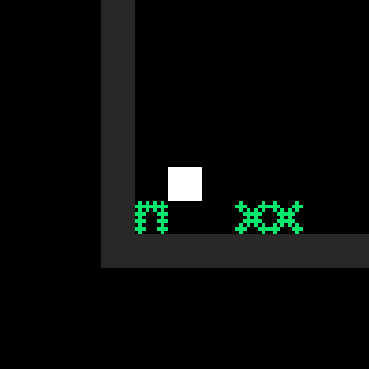}};
\path[arrow, line width=1mm] (pretrans.east) -- node[text width=3.5cm] {\Large Move to object,\\\phantom{blah}\\transform shape} (posttrans.west);
\node at (9, 0) (check) {\Large Check};
\path[arrow, line width=1mm] (posttrans.east) -- (check.west);

\begin{scope}[shift={(0,-7)}]
\node[align=left,text width=8cm] at (-1, 3) {\Huge \bf Exploit trial};

\node at (-3, 0) (init) {\includegraphics[width=4cm]{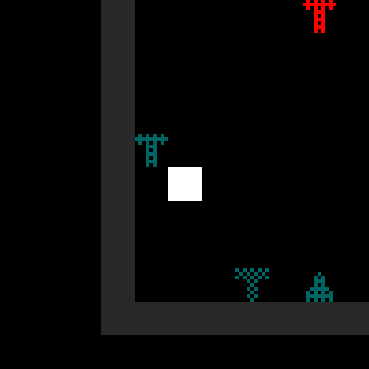}};
\node at (5, 0) (finalcheck) {\Large Choose correct object};
\path[arrow, line width=1mm] (init.east) -- (finalcheck.west);
\end{scope}
\end{tikzpicture}}
\caption{Environment \& task structure}\label{fig:ooo:env}
\end{subfigure}%
\begin{subfigure}[b]{0.33\textwidth}
\centering
\includegraphics[width=\textwidth]{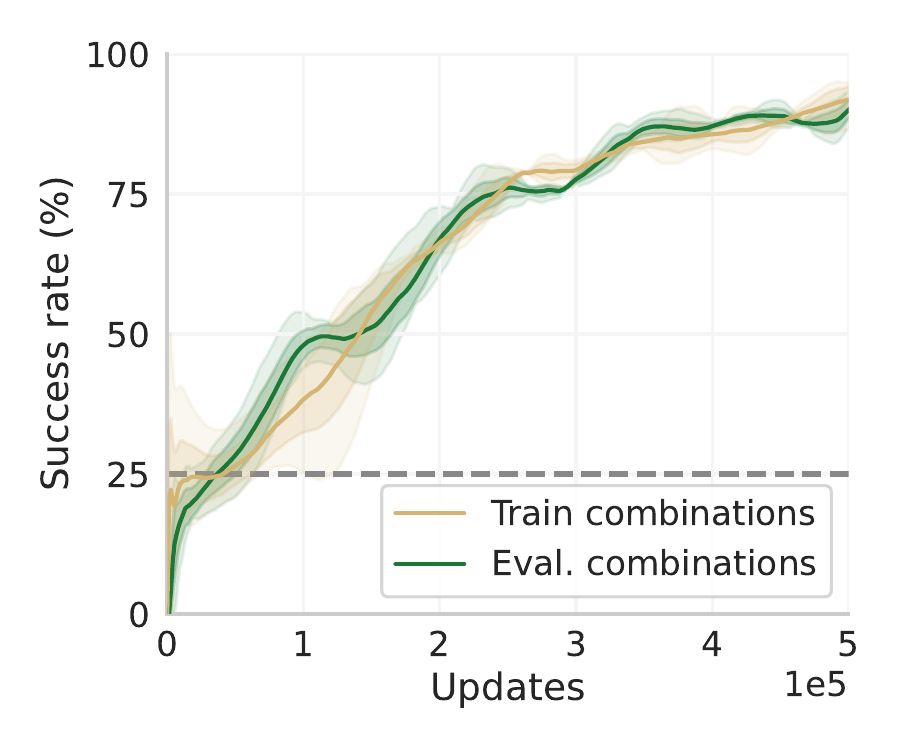}
\caption{Feature set generalization}\label{fig:ooo:feature_split}
\end{subfigure}
\begin{subfigure}[b]{0.33\textwidth}
\centering
\includegraphics[width=\textwidth]{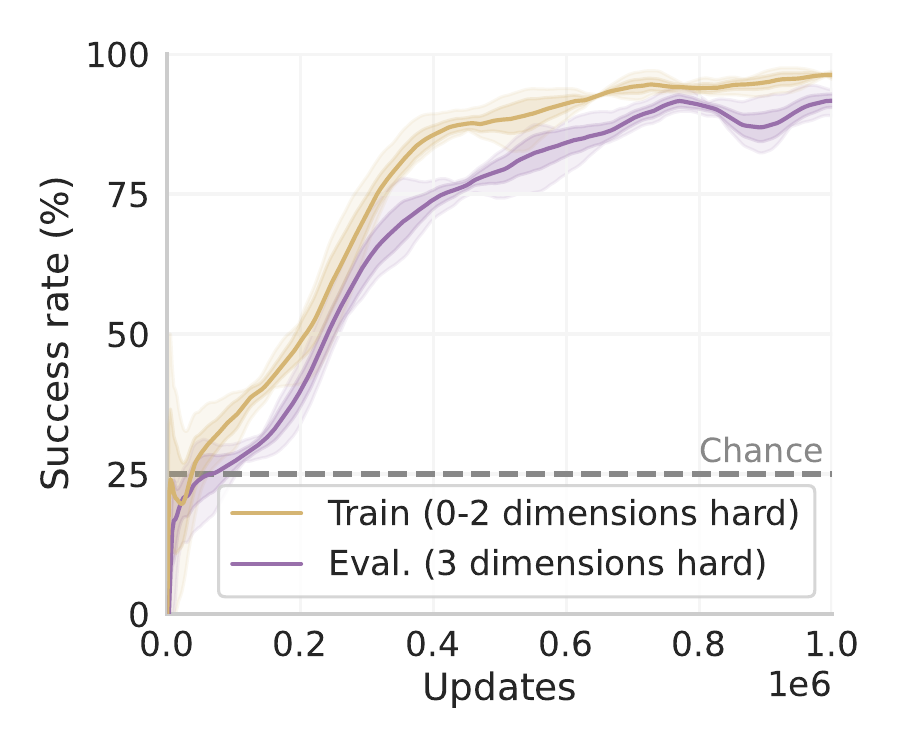}
\caption{Hard dimension generalization}\label{fig:ooo:hard_init_split}
\end{subfigure}
\caption{The odd-one-out meta-learning environments, where the goal is to pick up an object that is unique along the ``correct'' dimension, which cannot be directly observed. (\subref{fig:ooo:env}) An example of the observations and task structure. At top is an example experimentation trial. The agent is the white square (the view is egocentric, i.e. the agent is always in the center); it transforms one object to a unique shape, then it will ``pick up'' that object to check whether shape is the relevant dimension for this episode. After three experimentation trials, the agent proceeds to an exploit trial (bottom), where it can no longer transform objects, but where each object is unique along a different dimension; it must pick up an object according to the latent causal dimension discovered through experimentation. In training, the environment provides explanations of the objects and the results of experiments, which support learning. (\subref{fig:ooo:feature_split}) Generalization to novel feature combinations. (\subref{fig:ooo:hard_init_split}) Generalization from training data where 0-2 dimensions have a hard initialization (see text) to the case when all 3 dimensions do.  (Panel (\subref{fig:ooo:env}) adapted with permission from \citet{lampinen2022tell}. 3 seeds per condition.)} 
\label{fig:ooo}
\end{figure}
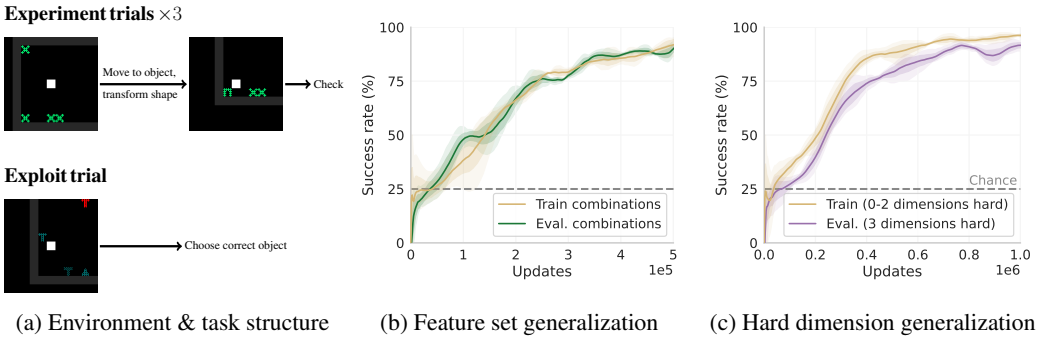
The above experiments used simple environments, where the agent directly observes the variables and the outcomes of its interventions. Here, we extend to a more complex set of environments involving a causal structure that can only be understood by interpreting relations among objects in high-dimensional pixel observations. Specifically, we experiment on the causal intervention/meta-learning versions of the odd-one-out environments \citep{lampinen2022tell}. See Fig. \ref{fig:ooo:env} for some example observations. 

In these environments, objects vary along multiple dimensions (color, shape, and texture). In each episode there is one dimension that is ``correct'' --- this ``correct'' dimension is a latent causal factor that the agent cannot directly observe. The agent is rewarded for picking up an object that is unique (the odd one out) along the ``correct'' feature dimension. The agent is first given a series of trials in which to experiment and determine which dimension is currently ``correct.'' Specifically, the agent is given a magic wand that allows it to transform features of objects, thus allowing it to make an object unique along one feature dimension. It can then pick up this object, and it can infer from whether it receives reward whether the dimension it tried is correct. These experiments allow the agent to succeed on a final exploit trial, where it no longer has access to a magic wand, but instead is faced with multiple objects that are each unique along a different dimension. To achieve a reward, the agent must use the results of its earlier experiments to choose the object that is unique in the correct way.

We evaluated two types of generalization. First, we simply split possible feature combinations into train and test sets. We evaluate whether the agent can learn, from passive data, an experimentation and exploitation strategy that generalizes to novel combinations. In Fig. \ref{fig:ooo:feature_split}, we show that the agent is able to learn these tasks well, and to generalize equally well to the evaluation conditions. Note that the tasks are designed such that simpler strategies --- such as relying on feature values --- will not yield above chance performance.

We then evaluated a more challenging generalization condition. The original paper \citep{lampinen2022tell} considered harder object initializations that make the experiments more difficult, by forcing the agent to transform one object, and then choose a \emph{different} object that is made unique by that transformation. We create a more challenging relational generalization split by training the agent in settings where 0-2 of the feature dimensions (randomly sampled in each training episode in the dataset) have this harder initialization. We then evaluate the agent's ability to generalize to evaluation in a harder setting where all 3 feature dimensions have this difficult initialization. This condition tests the agent's generalization to settings more challenging than those observed in the data. In Fig. \ref{fig:ooo:hard_init_split}, we show that the agent is able to learn and generalize these more difficult tasks.

\textbf{Explanations support learning:} In these experiments, we followed the method of \citet{lampinen2022tell}; the agent is trained to predict natural language explanations (provided by the environment during training) as an auxiliary loss. For example, when the agent performs a successful experiment that identifies the correct dimension as shape, during training this will be accompanied by an explanation like ``correct, the latent feature is shape, and this object is uniquely square.'' In Supp. Fig. \ref{app:fig:explanations:agent} we show that these explanations are beneficial even when learning from expert data; without predicting explanations, the agents do not learn or generalize these tasks as readily. Furthermore, explanations can enable other kinds of generalization from passive learning, as we explore in the next section.

\subsubsection{Explanations can deconfound for passive learners too}
\begin{wrapfigure}{r}{0.4\textwidth}
\vspace{-1.25em}
\includegraphics[width=0.35\textwidth]{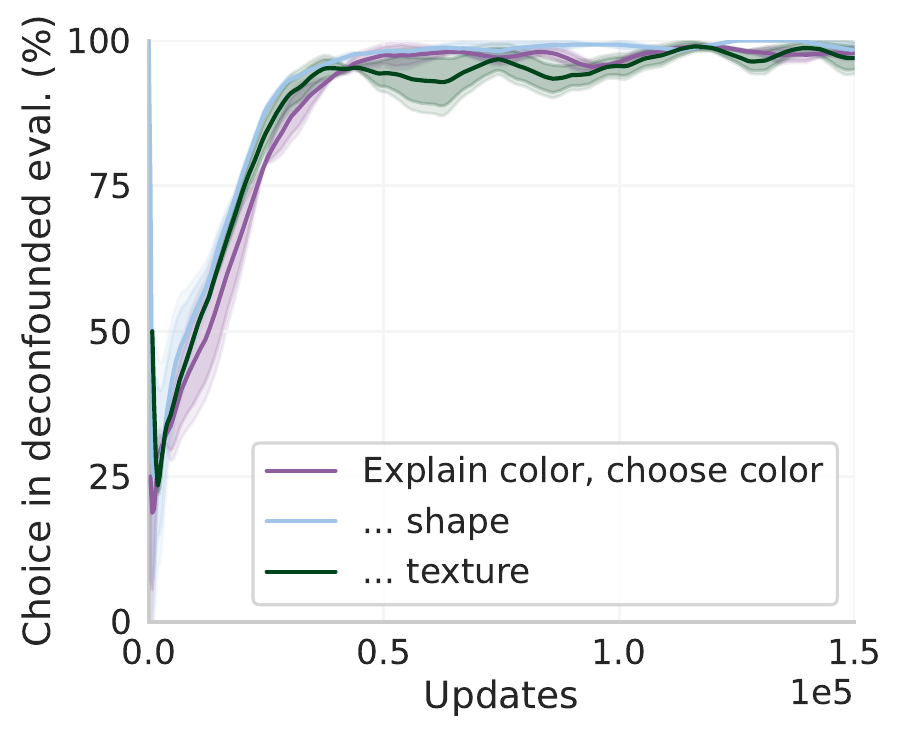}
\vspace{-0.25em}
\caption{Explanations can deconfound for passive learners trained on confounded data --- the agents generalize OOD in accordance with the explanations they were trained to predict.} 
\label{fig:ooo:deconfound}
\vspace{-1em}
\end{wrapfigure}

The original odd-one-out work also explored a setting of learning from perfectly confounded experiences, in which a rewarding object is unique along three feature dimensions, so it is unclear which feature(s) are relevant. The authors showed that RL agents can generalize out of distribution to follow any particular dimension in a deconfounded test (where different objects are unique along each dimension), if they are trained to predict language explanations that focus on that particular feature dimension. Although these experiments do not directly involve causal intervention, they are relevant to the broad question of what can be learned from confounded data --- and are particularly relevant to LMs, because explanations are a common feature of the human language from which these models learn. However, the prior work only experimented with active RL agents. Can passive learners also generalize out-of-distribution from predicting explanations?

We reproduced these deconfounding experiments in our passive learning setup, using the same expert as above. We find that passive learners also generalize consistently OOD according to the explanations they predicted during training (Fig. \ref{fig:ooo:deconfound}). Passive learners that were trained to predict explanations about shape would tend to choose the uniquely-shaped object in the deconfounded evaluation, and agents trained to predict explanations about texture or shape would analogously generalize according to those features. Thus, natural language explanations encountered during passive training can shape the way a system generalizes out-of-distribution at test time.

\subsection{Pretrained language models}

Finally, we tested whether pretrained language models could generalize causal strategies from a few-shot prompt. To do so, we adapt the odd-one-out intervention experiments from above to a language setting, and evaluate whether a large language model \citep[70B parameter Chinchilla, from][]{hoffmann2022training} can generalize a causal strategy in this environment from a few-shot prompt.
Specifically, we give the LM natural language observations of the objects in a scene, and pose choices between actions described in natural language. For the experimentation trials, we give a choice of which object to intervene on and which feature to transform. In the final trial, we simply give a choice among the objects, each of which is unique along one dimension. 

We give the LM four example shots, in all of which the latent ``correct'' dimension is sampled from only two of the three possible dimensions (color, shape, and texture), and evaluate the ability of the LM to generalize to an episode where the relevant dimension is the third, held-out one.
Note that this is a more difficult generalization setting than was investigated with the agents --- generalizing a causal relation to a dimension for which it was never observed in training, from only a handful of examples of other dimensions. We also experiment with holding out each of the dimensions, to ensure that the results are robust.
\begin{wrapfigure}{r}{0.55\textwidth}
\vspace{-0.25em}
\includegraphics[width=0.55\textwidth]{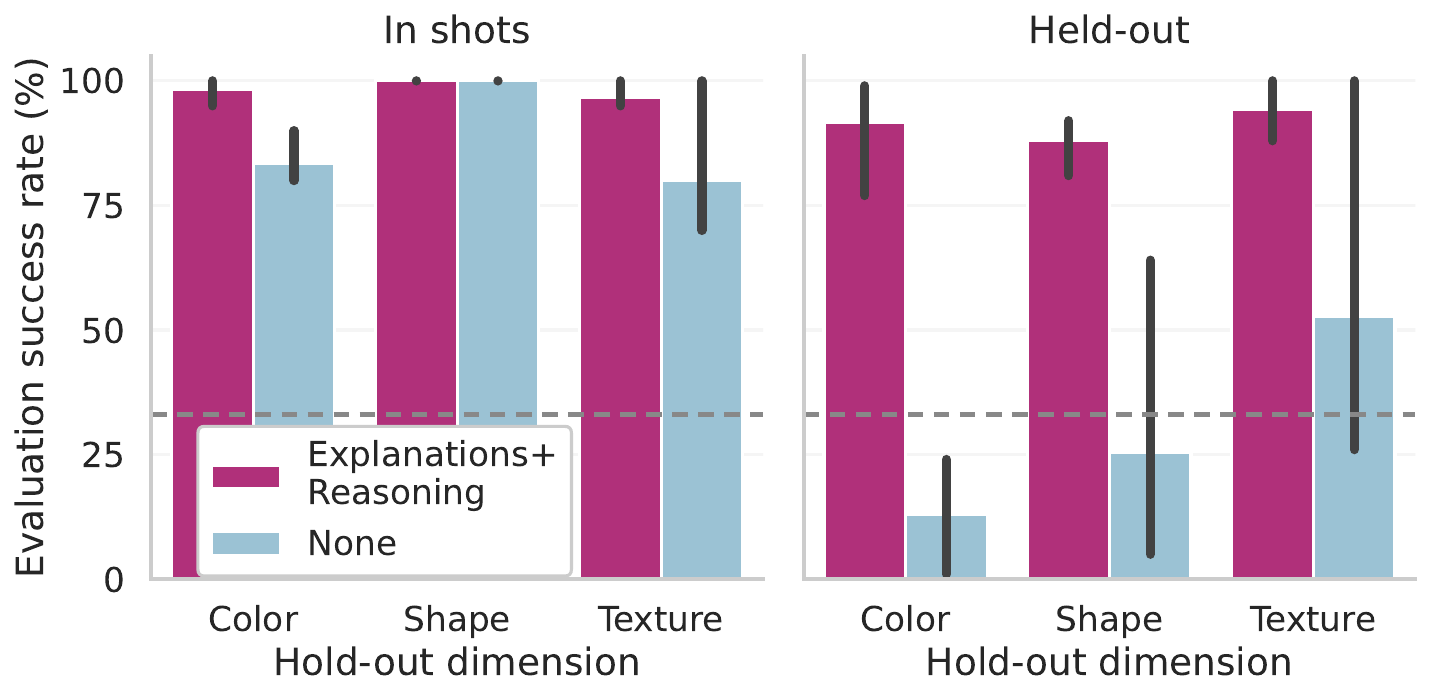}
\caption{LMs can generalize the odd-one-out tasks to new dimensions that were not correct in the example shots (right panel), but only if explanations/reasoning are included in the prompt. On new samples of the dimensions included in the examples (left), LMs can achieve some performance with or without explanations. (3 prompts per condition; errorbars are 95\%-CIs.)} 
\label{fig:lm}
\vspace{-0.5em}
\end{wrapfigure}
LM performance can be increased by incorporating factors like reasoning traces/chain-of-thought or explanations \citep{nye2021show,wei2022chain,suzgun2022challenging,zhou2022teaching,lampinen2022can,kojima2022large} in the prompt, and these strategies are particularly useful in challenging reasoning cases. We therefore include both reasoning traces and explanations in our prompt examples; at test time we allow the LM to generate its own explanations and reasoning traces at the appropriate places in the episode.
We also choose the example shots for the prompt by optimizing for performance on new samples of tasks involving the two trained dimensions \citep{liu2023pretrain,lampinen2022can}. For more details, including an example episode, see Appx. \ref{appx:envs:ooo:lm}.

In Fig. \ref{fig:lm} we show the results. Given a strong four-shot prompt, the language  model can generalize correctly to new instances of the dimensions that were relevant in the prompt; critically, it can also generalize to new instances of the held-out dimension in each case (and even to a somewhat harder test with an added object; Appendix \ref{app:supp_exp:lm_extra_object}.) However, explanations are critical for this generalization. In Appendix \ref{app:supp_exp:lm_all_explanations} we show that either explanations or chain-of-thought reasoning suffice to provide some benefit, but results are most reliable when both are provided together. Thus, the language model is able to build on its strong prior knowledge to generalize, but only when supported by some form of explanation.

\section{Discussion}

What can be learned from passive data about strategies for experimenting to discover and exploit causal structure? Our work illustrates that a non-trivial amount can, at least when the passive data contains the results of experiments. From passive imitation, agents can learn strategies for experimentation that can be generalized to correctly experiment on novel causal structures, and learn strategies for exploiting the results of their own experiments to achieve goals under those novel structures. Agents can even generalize in complex environments where they receive high-dimensional observations; likewise, passively-trained language models can generalize from a high-quality few-shot prompt. Furthermore, our results highlight the role that natural language explanations can play in drawing out latent causal structure to enable more generalizable learning. Explanations can even enable passive learners to generalize nearly perfectly from otherwise perfectly-confounded data. Passive learning can be surprisingly powerful under the right circumstances.

These observations provide perspective on the kinds of understanding of causality and experiments that can be acquired by language models. For example, consider the following, from \citet{pearl2018theoretical}: 
\begin{quote}
 ``This hierarchy, and the formal restrictions it entails, explains why statistics-based machine learning systems are prevented from reasoning about actions, experiments and explanations. [...] the hierarchy denegrades [\emph{sic}] the impressive achievements of deep
learning to the level of Association [...] Unfortunately, the theoretical barriers that separate the three layers in the hierarchy tell us that the [...] objective function does not matter. As long as our system optimizes some property of the observed data, however noble or sophisticated, while making no reference to the world outside the data, we are back to level-1 of the hierarchy with all the limitations that this level entails.''
\end{quote}
Naively, it might seem that these arguments doom language models to never acquire real knowledge of causality \citep[cf.][]{willig2023causal}. 
Our results provide useful formal and empirical nuance. While there are certainly restrictions to learning from passive data (see below for further discussion), it is possible to learn generalizable strategies for causal experimentation and intervention from passive data alone --- at least if the data include examples of an expert intervening, and perhaps explanations.

Other work has engaged with these issues. In particular, \citet{ortega2021shaking} provide a theoretical account of the effects of confounding on passively trained sequence models, and propose a counterfactual training method to enable an adaptive system to learn effective intervention strategies in the presence of confounders. However, this method relies on having active access to an expert teacher rather than simply offline data; the authors note that ``in general, we cannot meta-train an agent only from expert demonstrations to imitate the expert at deployment time'' and ``[t]o address the problem, it is necessary to incorporate assumptions (e.g. a set of hypotheses) about the confounding variable, but currently it is unclear how to incorporate this seamlessly into a meta-learning scheme.'' Our work complements theirs, by illustrating some of the cases in which an agent can successfully generalize causal experimentation and goal-seeking strategies from passive data alone --- even in the presence of confounders --- and by highlighting the role that natural language explanation can play in incorporating expert knowledge to allow active generalization from passive learning.

\textbf{Explanations \& causal structure:} Explanations were important for effective learning or generalization in the odd-one-out tasks; explanations can even allow generalizing from otherwise totally-confounded data. These results fit with prior work showing that explanations and reasoning traces help agents \citep{lampinen2022tell,nguyen2021interactive} and LMs \citep{narang2020wt5,hase2021can,wei2022chain,lampinen2022can,hsieh2023distilling}. More broadly, explanations are key for human learning because they explicitly highlight generalizable, causal structures \citep{lombrozo2006functional,lombrozo2006structure,kirfel2022inference}. For example, explanations can constrain the hypothesis space of causal graphs that a system needs to consider \citep[cf.][]{nam2023show}. Alternatively, language\footnote{Or other forms of abstraction.} may obviate the need for explicit causal graphs. That is, explanations could serve as the ``reference to the world outside the data'' that \citet{pearl2018theoretical} sought in the quote above.

\textbf{Language models:} Our experiments have obvious implications for the causal abilities of language models. Prior works have suggested that LMs can only parrot causal structures observed in training \citep{willig2023causal}. However, our experiments illustrate that this should not be assumed; the implications of passive learning for causal strategies can be subtle. Passive data is not necessarily strictly observational. In particular, observing other's interventions, especially with explanations, can enable generalization. LM training data will contain many examples describing interventions and outcomes together with explanations thereof --- whether scientific articles, or just descriptions of a debugging process. Just as the data in a scientific paper do not become observational if we read the paper rather than writing it, the data on the internet do not become observational just because an LM consumes it passively. Thus, sequences of interventions, outcomes, and explanations may in some cases be sufficient to allow an LM to generalize to proposing experiments and using the results to draw appropriate conclusions, at least if given an appropriate prompt.

\textbf{Tool use \& agency:} These findings are of particular interest given the recent focus on language models using tools \citep[e.g.][]{schick2023toolformer}. Tool use is a clear case of active interaction in the world, which one might assume requires some knowledge of causality. Our results may help explain the successes of these methods with models trained purely from passive supervised learning (though note that many applications of LMs are also actively fine-tuned, which may also contribute, see below). 
A related question is whether LMs, despite passive learning, can acquire an ability to act as goal-directed agents. \citet{andreas2022language} notes that effective language modeling requires modeling the goals of the communicator that produced the language, and thus suggests that through this process LMs learn to at least \emph{model} some aspects of agency. Our results may provide some support for these arguments, as generalizing goal-directed behavior to new causal structures is a key aspect of agency. 

\textbf{Experimental limitations:} Our experiments are limited in a number of important ways. First, the causal structures we used are small, finite acyclic graphs. While this limitation (i.e., that the causal graph is a DAG over a handful of variables) is assumed by much of the prior causal literature, many real world causal structures are much more complex. For example, they may include cycles, at least over time.\footnote{Though these can be accommodated in a DAG by ``unrolling'' the graph.} Changing these assumptions changes the nature of the causal discovery problem \citep{bongers2021foundations,forre2018constraint}; thus it is an open question whether agents could similarly learn causal strategies in these more complex settings. More generally, even our most complex environments have relatively simple and constrained observation and action spaces compared to the real world, and the space of possible experiments is generally small and heavily constrained. Our agents are also limited to fit with their environments (e.g. having the right number of observation inputs and action outputs for the number of variables in the DAG environment). It is an open question whether agents could in practice learn causal strategies from passive data in environments that relaxed these simplifying assumptions, though our results suggest it would be possible in principle.

\textbf{What this paper does \emph{not} imply:} We do not intend to suggest that passive learning is as beneficial as active learning, or that confounding is not a challenge for passive learners. Active learning can be substantially more effective or efficient than passive learning in humans \citep{gureckis2012self,markant2014better}, animals \citep{held1963movement}, and RL agents \citep{oudeyer2007intrinsic,bohg2017interactive,ostrovski2021difficulty,carta2023grounding}. It can be difficult to learn from purely positive examples, without some active correction of off-optimal-policy behavior \citep[e.g.][]{ross2011reduction,dehaan2019causal}. Indeed, while the LM we evaluated was only trained passively, many deployments of LMs incorporate interactive learning through RL from Human Feedback [\citealp{christiano2017deep}; e.g.\ \citealp{ouyang2022training}]; one key benefit of this may be the value of training the model on active, on-policy data. We therefore expect that agents or LMs that are fine-tuned in an active setting, or with more sophisticated causal corrections \citep{ortega2021shaking}, will generally develop more robust and generalizable causal strategies than those trained in purely passive settings. 

Furthermore, confounding is clearly a challenge for any system that attempts to discover and use causal structure. The ongoing debates about confounding in the medical literature \citep[e.g.][]{krittanawong2022alcohol,zhao2023association} illustrate that this is a challenge even for causal experts such as human scientists. However, our work contributes to understanding the regimes in which confounding during passive learning can be overcome by experimentation at test time. We consider a range of confounds; even our train-test splits effectively correspond to strong confounds which are not directly observable, and for which the agent/LM only experiences training data sampled under one value, and then is tested on situations sampled under another. Nevertheless, agents can in some cases generalize despite these confounds, particularly when supported by explanations. These results highlight the role that language and explicit descriptions of the quest for causal structure can play in acquiring causal strategies. 

\textbf{Conclusions:} We illustrate the potential for purely passive learning of active causal strategies across various settings, particularly with the support of explanations. These results have implications for understanding the causal limitations of passive learning for a system that can only interact at test time, and have particular relevance to understanding the capabilities and behaviour of language models.

\subsection*{Acknowledgements}
We thank Nan Rosemary Ke, Tobias Gerstenberg, and the anonymous reviewers for helpful comments and suggestions.

\bibliography{neurips_2023}
\bibliographystyle{plainnat}

\clearpage
\appendix

\section{Environments} \label{appx:envs}

\subsection{DAG environment} \label{appx:envs:dag}

The causal structure is sampled according to the following process:

\begin{enumerate}
    \item An ordering of the variables is picked at random. (At test time, the test goal variable is constrained to be at the end of the order, but the order is otherwise unconstrained.)
    \item A noise variance for each of the variables is sampled, from a (clipped) normal distribution with mean 0.5 and standard deviation 0.25.
    \item (In the adaptive experimentation version from Section \ref{sec:experiments:dag:adaptive}, a subset of half the variables are sampled to be relevant; all subsequent steps only include variables from this subset.)
    \item A number of independent variables \(i\) is sampled, and the first \(i\) variables in the ordering are set to be independent.
    \item For the remaining nodes, their input weights are sampled as follows:
      \begin{itemize}
          \item 1 or 2 nodes from earlier in the order are sampled to be parents of this node. (In training, the test goal variable is constrained to not have parents that are descendents of the test intervention target. At test time, it is selected to \emph{only} have parents that are descendants of the test intervention target.)
          \item The input weight from each parent is sampled uniformly in \([-2, 2]\). 
          \item At test time, all weights on the path from the intervention target to the goal are increased in magnitude to larger than 1, to ensure that intervening on the target (or its ancestors, depending on the test condition) is the optimal way to maximize the goal node. All weights from other inputs to any nodes along the path are decreased to less than 1.
      \end{itemize}
\end{enumerate}

These constraints allow a total of 1,277 possible DAG structures for training.

When propagating values through the graph, each node's value is computed as a linear function of its ancestors outputs, followed by the addition of a value sampled from a normal distribution with mean 0 and the current noise variance of this node, finally followed by a nonlinearity. We used a leaky relu nonlinearity \citep{maas2013rectifier} with a leak current of 0.2; we also compare to the case without a nonlinearity in Appendix \ref{app:supp_exp:linear_dag}.

\textbf{Observations:} The environment observations contain the following features:
\begin{itemize}
    \item The values of the variables before the previous intervention.
    \item The previous intervention taken (as a numerical value of the intervention on the variable it was applied to, e.g. \([0, 0, -4, 0, 0]\)).
    \item The outcome values after the previous intervention.
    \item The initial values for the next experiment.
    \item The goal variable to maximize, as a one-hot value (or zeros before the goal is presented).
    \item (Only in the adaptive experimentation conditions, a several-hot vector indicating which variables are relevant.)
\end{itemize}

The environment offers a discrete action space of size \(2 \times n\), corresponding to the two possible interventions (\(-4\) or 4) that the agent can perform on any of the \(n\) variables.

\textbf{Expert policy:} In the exploration phase, the expert policy randomly chooses a variable to intervene on first, then intervenes on the variables in order after that, looping back to 0 after it reaches the end; e.g., the expert intervention order might be [2, 3, 4, 0, 1]. Whether the expert interventions are positive or negative is chosen randomly on each step. In the exploitation phase, the expert takes the optimal action to increase the value of the goal variable.

\textbf{Heuristic baseline policies:} We evaluate the baselines only during the goal-seeking/exploitation phase. The value baseline tracks the outcomes of interventions during the exploration phase, then at exploitation time takes the action which gave the highest value on the target node. The change baseline is similar, but sign sensitive; it chooses the intervention that changed the value of the goal node the most, and then flips the sign of the intervention if the original change in the outcome was negative.

\textbf{Correlation baseline policies:} The correlation baselines constructs a policy based on the correlational statistics of the observations the agent makes during training; by fitting linear regression models. WLOG, assume the target node is E, and there are \(n = 5\) nodes. The total correlation baseline fits \(n-1\) separate univariate regressions, one from A to E, one from B to E, etc. The partial correlation baseline fits a single multivariate regression from \(A, B, C, D \rightarrow E\); that is, it controls for the effect of the other variables. In either case, the variable with the largest regression slope is chosen to intervene on, with a sign corresponding to the sign of the slope. 

\subsection{Odd-one-out environments} \label{appx:envs:ooo}

We use the open-sourced version of the odd-one-out environments \citep{lampinen2022tell}, which is available at: \url{https://github.com/deepmind/tell_me_why_explanations_rl}. 

\textbf{Feature combination split:} We simply split combinations of features randomly into a train and test set by their hash value; we hold out 20\% of the hash values for testing. This is a simple split, so it is not particularly surprising that the agent generalization performance is comparable to train performance.

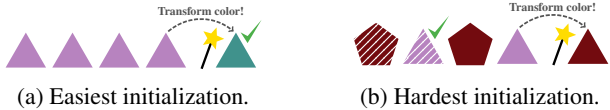
\begin{figure}[H]
\centering
\begin{subfigure}[b]{0.33\textwidth}
\centering
\resizebox{0.75\textwidth}{!}{
\begin{tikzpicture}
\node[fill=bpurp!70!white,regular polygon, regular polygon sides=3,  minimum height=2cm] at (0, -0.25) {};
\node[fill=bpurp!70!white,regular polygon, regular polygon sides=3,  minimum height=2cm] at (2, -0.25) {};
\node[fill=bpurp!70!white,regular polygon, regular polygon sides=3,  minimum height=2cm] at (4, -0.25) {};
\node[fill=bpurp!70!white,regular polygon, regular polygon sides=3,  minimum height=2cm] at (-2, -0.25) {};

\path[draw, line width=1mm] (5.53, -0.8) to (6, 0.5);
\node[star, star points=5, minimum size=1 cm, star point ratio=2.25, fill=yellow!80!borange, rotate=-20] at (6, 0.5) {};

\path[arrow, dash pattern=on5off2, darkgray!90!white, line width=0.75mm, bend left=40] (4.1, 0.8) to node[yshift=10] {\Large \bf Transform color!} (6.9, 0.8);
\node[fill=bgreen!40!bblue,regular polygon, regular polygon sides=3,  minimum height=2cm] at (7, -0.25) {};

\path[fill=bgreen] (7.2, 0.76) -- (7.5,0.3) -- (8.1, 1.3) -- (7.5,0.6) -- cycle;
\end{tikzpicture}
}
\captionsetup{width=.8\textwidth}
\caption{Easiest initialization.}  \label{app:fig:meta_relations:easy}
\end{subfigure}%
\begin{subfigure}[b]{0.33\textwidth}
\centering
\resizebox{0.75\textwidth}{!}{
\begin{tikzpicture}
\node[fill=bpurp!70!white,regular polygon, regular polygon sides=3,  minimum height=2cm] at (0, -0.25) {};
\node[fill=bred!70!black,regular polygon, regular polygon sides=5,  minimum height=2cm] at (2, 0) {};
\node[fill=bpurp!70!white,regular polygon, regular polygon sides=3,  minimum height=2cm] at (4, -0.25) {};
\node[fill=bred!70!black,regular polygon, regular polygon sides=5,  minimum height=2cm] at (-2, 0) {};
\draw[draw=white, pattern=north east lines, pattern color=white] (-3, -1) rectangle (1,1);

\path[draw, line width=1mm] (5.53, -0.8) to (6, 0.5);
\node[star, star points=5, minimum size=1 cm, star point ratio=2.25, fill=yellow!80!borange, rotate=-20] at (6, 0.5) {};

\path[arrow, dash pattern=on5off2, darkgray!90!white, line width=0.75mm, bend left=40] (4.1, 0.8) to node[yshift=10] {\Large \bf Transform color!} (6.9, 0.8);
\node[fill=bred!70!black,regular polygon, regular polygon sides=3,  minimum height=2cm] at (7, -0.25) {};
\path[fill=bgreen] (0.2, 0.76) -- (0.5,0.3) -- (1.1, 1.3) -- (0.5,0.6) -- cycle;
\end{tikzpicture}
}
\captionsetup{width=.8\textwidth}
\caption{Hardest initialization.}  \label{app:fig:meta_relations:hard}
\end{subfigure}
\caption{Cartoon examples of possible object sets from the (\subref{app:fig:meta_relations:easy}) easiest train, and the (\subref{app:fig:meta_relations:hard}) hard test level in the odd-one-out dimension split tasks. In the simplest case, the agent can choose the object it has just transformed. However, in the harder levels, the agent has to choose a \emph{different} object, which is left unique after its transformation, in order to test a dimension. Choosing the object it transformed will always be incorrect. Note that these illustrations are just cartoons; the agent still saw pixel observations like those shown in Fig. \ref{fig:ooo:env}. (Adapted with permission from \citet{lampinen2022tell}.)} \label{app:fig:meta_relations}
\end{figure}

\textbf{Hard dimension split:} The odd-one-out environments afford several possible initialization of the objects before the experiments, that affect how hard the experiments are to perform (Fig. \ref{app:fig:meta_relations}). We train the agent on the three easier initializations (with 0-2 dimensions having the hard setting), and test on the case where all three do. 

\textbf{Expert policy:} We augmented these environments with a hand-designed expert policy which first navigates towards the center of the room until objects come into view, then goes to the nearest object, transforms a feature of it, and then goes to the object which is made unique (not necessarily the transformed one), and chooses it. The expert policy experiments on the three possible feature dimensions (color, shape, and texture) in order. Once it has observed the correct dimension, it continues to choose that (rather than experimenting on the others). On the final trial, it navigates to the object which is unique along the correct dimension. The expert's navigation policy tries to move towards its navigation target as efficiently as possible, with simple one-step heuristics for navigating around obstacles in the way (first try navigating around in a way that brings you closer to your target, if possible; otherwise fall back to successively more roundabout routes).

\subsubsection{Language model adaptation of the odd-one-out intervention tasks} \label{appx:envs:ooo:lm}

In order to adapt the odd-one-out tasks to the language models, we converted the observations and actions into language. We also reduced the number of objects to three, used the simple contrast structure (all objects initialized the same in the experimenting episodes), and removed the locations and navigation (which are more difficult to describe, and do not contribute substantially to the conceptual issues). We provide a full example episode from the expert policy in Listing \ref{list_ooo_lm}.

\begin{lstlisting}[label=list_ooo_lm,caption=Example expert episode in the LM odd-one-out environment,float,breaklines]
New game:

There is a set of three objects in front of me:
A) white square striped
B) white square striped
C) white square striped

I transform object A into a different color: pink.
Choosing object A was not rewarded.
Explanation: The rewarding dimension must not be color.

There is a set of three objects in front of me:
A) black ellipse striped
B) black ellipse striped
C) black ellipse striped

I transform object B into a different shape: trapezoid.
Choosing object B was rewarded!
Explanation: In this game, I am rewarded for unique shape.

There is a set of three objects in front of me:
A) red pentagon solid
B) red pentagon solid
C) red pentagon solid

I transform object C into a different texture: striped.
Choosing object C was not rewarded.
Explanation: The rewarding dimension must not be texture.

There is a set of three objects in front of me:
A) purple ellipse solid
B) green trapezoid solid
C) green ellipse striped

Reasoning: Let's think step by step. In this game, I am rewarded for unique shape. Object B is the only trapezoid object, because A and C are ellipse, so B has a unique shape. I will be rewarded for choosing object B.
I choose object B
Choosing object B was rewarded!
Explanation: I was rewarded for unique shape in this game.
\end{lstlisting}

In an evaluation episode for the language model, it was given a four shot prompt of examples of this type (separated by lines of equals signs and ``New game:'' to delimit a new episode). It would first receive an observation of the first objects, and would be prompted with this followed by ``I transform object'' and given a choice between A, B and C. The object assigned the highest likelihood as a continuation would be chosen. Then, the prompt was continued through ``into a different'' and the model was given a choice between the different possible dimensions (color, shape, and texture). Again, the highest-likelihood continuation was chosen. The environment would then transform the object according to its rules, and the resulting attribute would be presented to the agent. Because we are using the simpler attribute relations, we assumed that the model chose the object it had just transformed. The model was then presented with the reward (or lack thereof) from this experiment.

After each trial, the model was prompted with ``Explanation:'' and allowed to generate until it produced a newline character. This was inserted into the prompt, followed by the beginning of the next trial. (For the few shot examples, we used explanations that connected the outcome of the experiment to the latent task structure, as illustrated above.)

On the final trial, the language model was presented with three distinct objects, each of which was unique along one of the three dimensions, in a random order. The LM was then prompted with ``Reasoning: Let's think step by step.'' \citep[cf.][]{kojima2022large} and again allowed to generate until a newline. (During training, we provided a fixed reasoning format, as illustrated above.) This reasoning trace was appended to the prompt, followed by ``I choose object'' and a choice between A, B, and C, which was then scored for whether it was correct along the relevant dimension.

\textbf{Expert policy:} We found that the expert policy affected generalization --- in particular, it was important for the expert to use a fixed experimentation policy (always experiment on first color, then shape, then texture); see Appendix \ref{app:supp_exp:lm_varied_expert}.

\textbf{Prompt construction:} We constructed the four-shot prompt for each condition by sampling 10 possible 4-shot prompts, then evaluating each of them on 20 fixed episodes sampled from the train distribution --- i.e. if texture dimension was held out, these 20 episodes would involve color or shape. The same 20 episodes were used for evaluating each of the 10 prompts. We selected the prompt that achieved the highest score on this validation set to use for evaluation in the hold-out condition.

\section{Supplemental experimental results} \label{app:supp_exp}

\subsection{The linear causal DAG case} \label{app:supp_exp:linear_dag}

We also removed the nonlinearities and evaluated the case of a fully-linear causal DAG, which makes the learning problem slightly simpler, and makes the correlational strategies stronger baselines. Indeed, if the noise variances on the nodes were fixed, the causal structure could be recovered from observational data alone in this special case \citep{peters2012identifiability}. Correspondingly, the baseline approaches match the optimal expert more closely in this setting. Nevertheless, the agent still matches the expert somewhat better than it matches the baselines, even in this case.

\begin{figure}[H]
\centering
\begin{subfigure}[b]{0.3\textwidth}
\centering
\includegraphics[width=\textwidth]{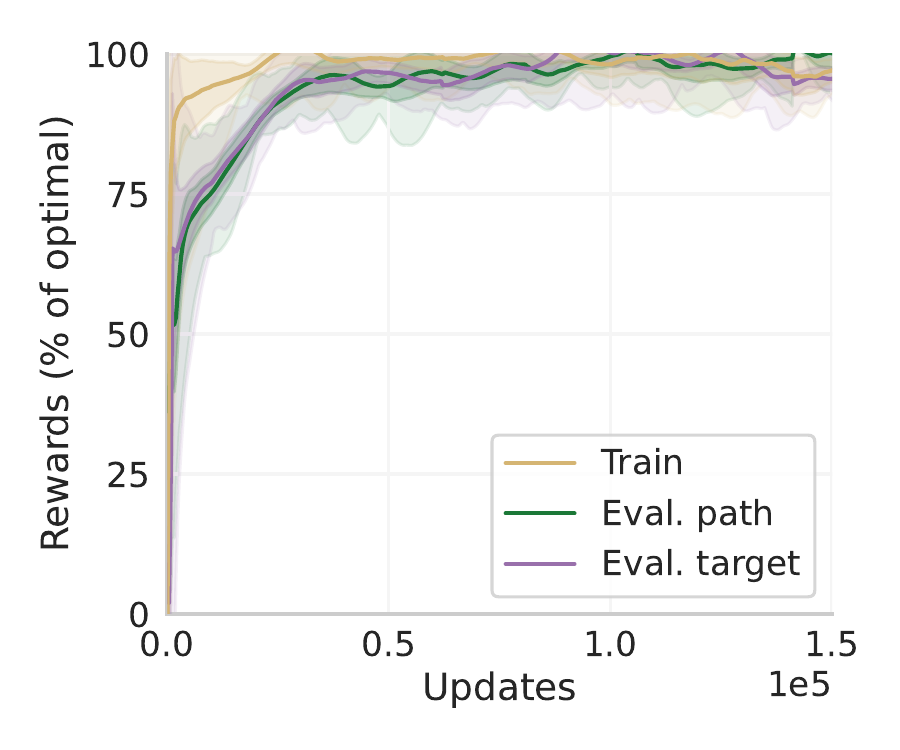}
\vskip-0.25em 
\caption{Rewards}\label{app:fig:linear_dag:rewards}
\end{subfigure}
\begin{subfigure}[b]{0.445\textwidth}
\centering
\includegraphics[width=\textwidth]{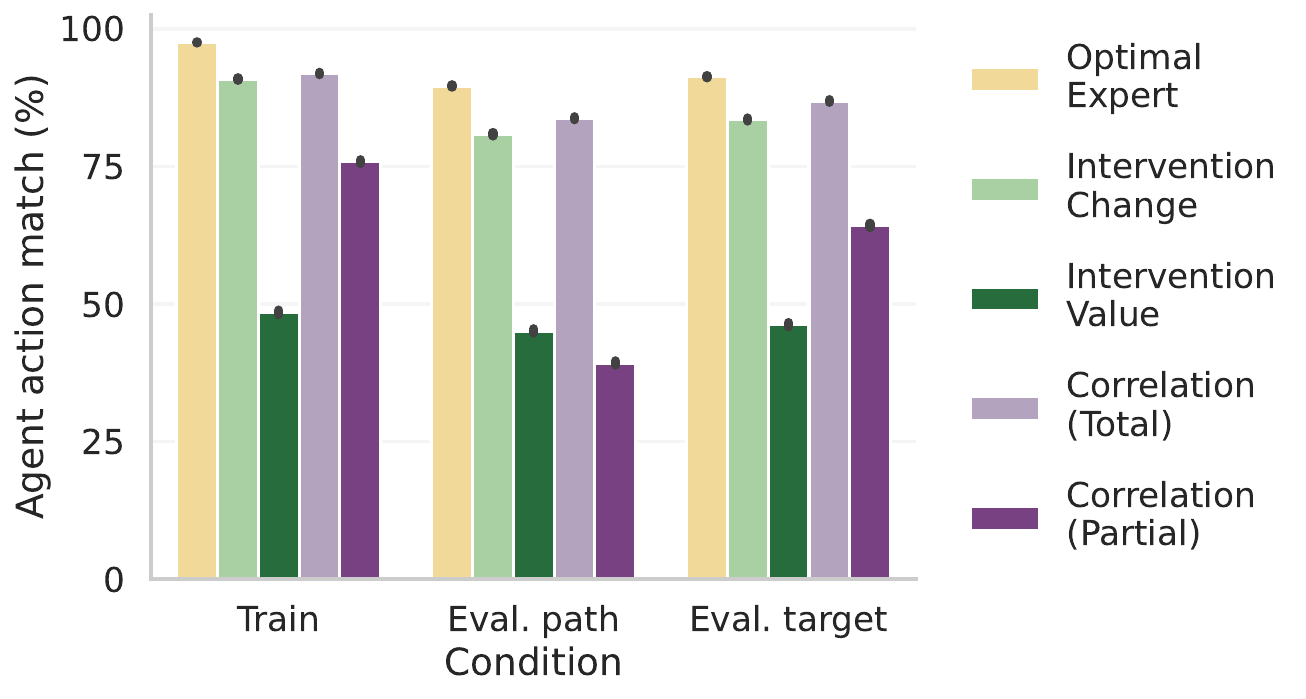}
\caption{Action analysis}\label{app:fig:linear_dag:actions}
\end{subfigure}
\caption{Experimental results in the linear version of the causal DAG environment. (\subref{app:fig:linear_dag:rewards}) Rewards obtained by the agent when evaluated in interactive settings, as a percentage of the optimal rewards. (\subref{app:fig:linear_dag:actions}) The proportion of the agent's actions that match the optimal behavior, or baselines based on heuristics from the interventions or correlational statistics. While the baselines are more effective in the fully-linear case, the agent still matches the expert strategy significantly more closely than it matches the heuristic baselines. (Error bars are 95\%-CIs.)} 
\label{app:fig:linear_dag}
\vskip-1em
\end{figure}

\subsection{The test-time exploration phase, with active intervention, is critical} \label{app:supp_exp:no_explore_ablation}

\begin{figure}[H]
\centering
\begin{subfigure}[b]{0.33\textwidth}
\centering
\includegraphics[width=\textwidth]{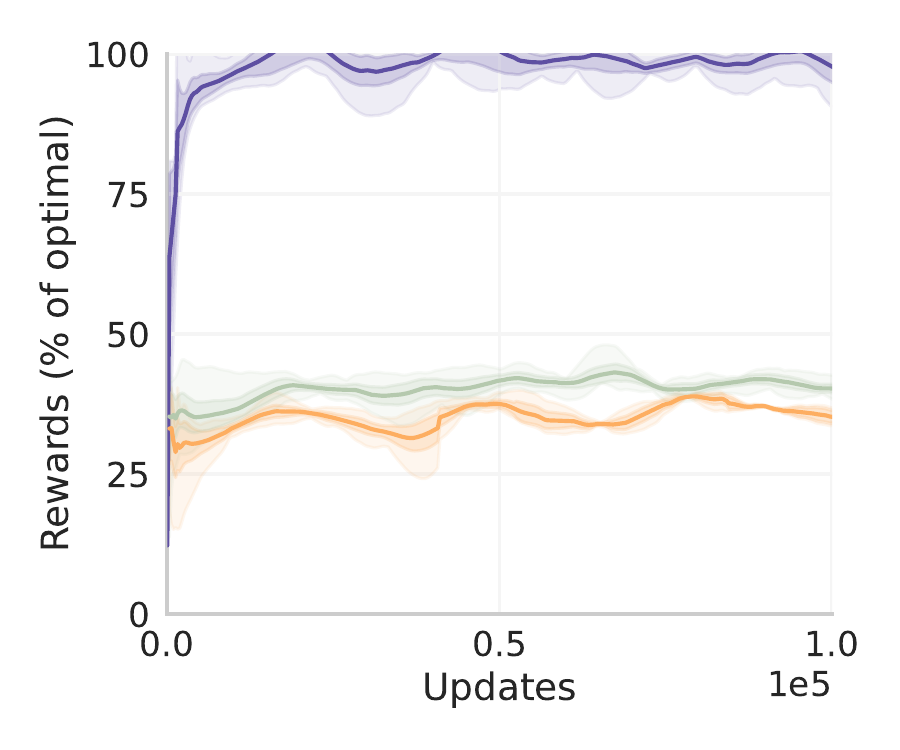}
\caption{Train.}\label{app:fig:explore_ablate:train}
\end{subfigure}%
\begin{subfigure}[b]{0.33\textwidth}
\centering
\includegraphics[width=\textwidth]{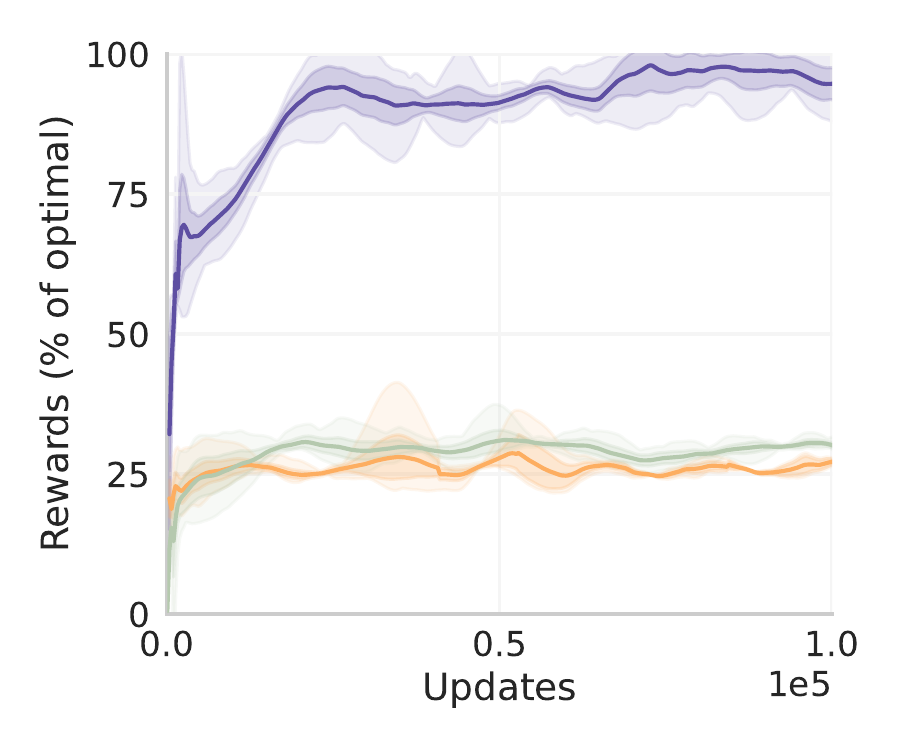}
\caption{Eval.\ path.}\label{app:fig:explore_ablate:eval_path}
\end{subfigure}%
\begin{subfigure}[b]{0.33\textwidth}
\centering
\includegraphics[width=\textwidth]{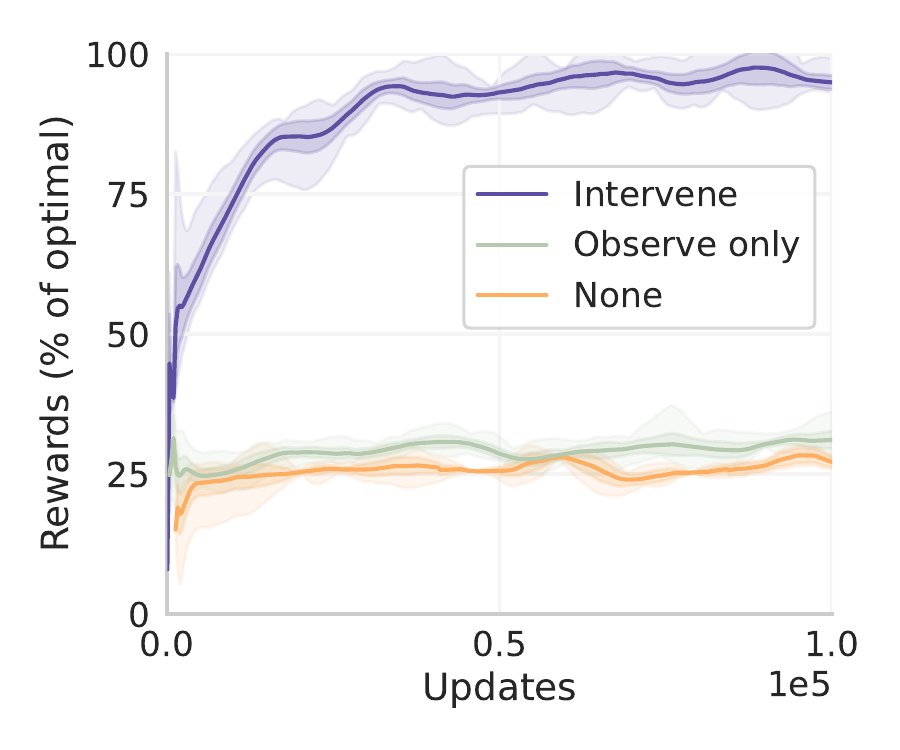}
\caption{Eval.\ target.}\label{app:fig:explore_ablate:eval_target}
\end{subfigure}%
\caption{Actively exploring at evaluation time is necessary for effective performance. In these ablation experiments, we remove the active (interventional) exploration phase, either by skipping it entirely (grey-green curves) or by replacing it with a purely observational phase, where the agent cannot intervene, but merely observes samples from the natural distribution of the causal DAG. Compared to agents that can intervene to explore (purple), as in the main experiments, the ablation conditions perform much worse on both (\subref{app:fig:explore_ablate:train}) graphs sampled from the training distribution and (\subref{app:fig:explore_ablate:eval_path}-\subref{app:fig:explore_ablate:eval_target}) graphs sampled from the evaluation distributions. Observing samples from the graph is better than nothing, but still results in much lower performance than intervening.} \label{app:fig:explore_ablate}
\end{figure}

The main text emphasizes the importance of agents intervening at test time; however, is this actually necessary? In these ablation experiments, we test this question by removing the active intervention exploration phase from both training and testing, either skipping it entirely or by replacing it with a purely observational phase, where the agent cannot intervene, but merely observes samples from the natural distribution of the causal DAG. Both ablation agents perform much worse than agents that can intervene (Fig. \ref{app:fig:explore_ablate}), thus demonstrating the importance of intervention at test time to determine the new causal structure.

\subsection{Generalization degrades if agents are trained on few DAGs structures} \label{app:supp_exp:more_holdouts}

\begin{figure}[H]
\centering
\begin{subfigure}[b]{0.5\textwidth}
\centering
\includegraphics[width=0.66\textwidth]{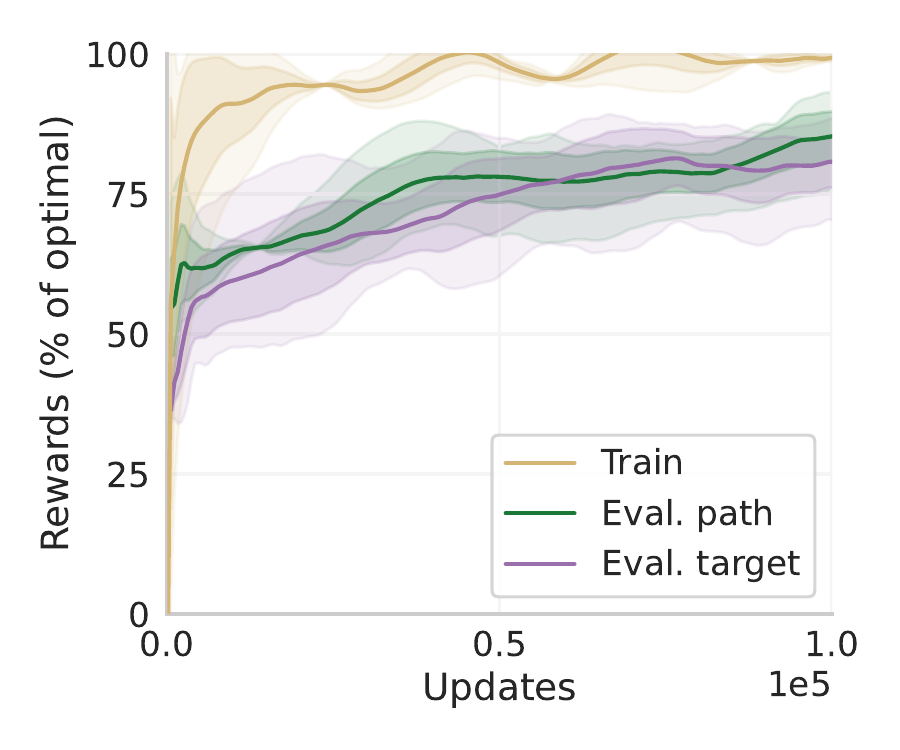}
\caption{428 training DAG structures.}\label{app:fig:more_holdouts:more}
\end{subfigure}%
\begin{subfigure}[b]{0.5\textwidth}
\centering
\includegraphics[width=0.66\textwidth]{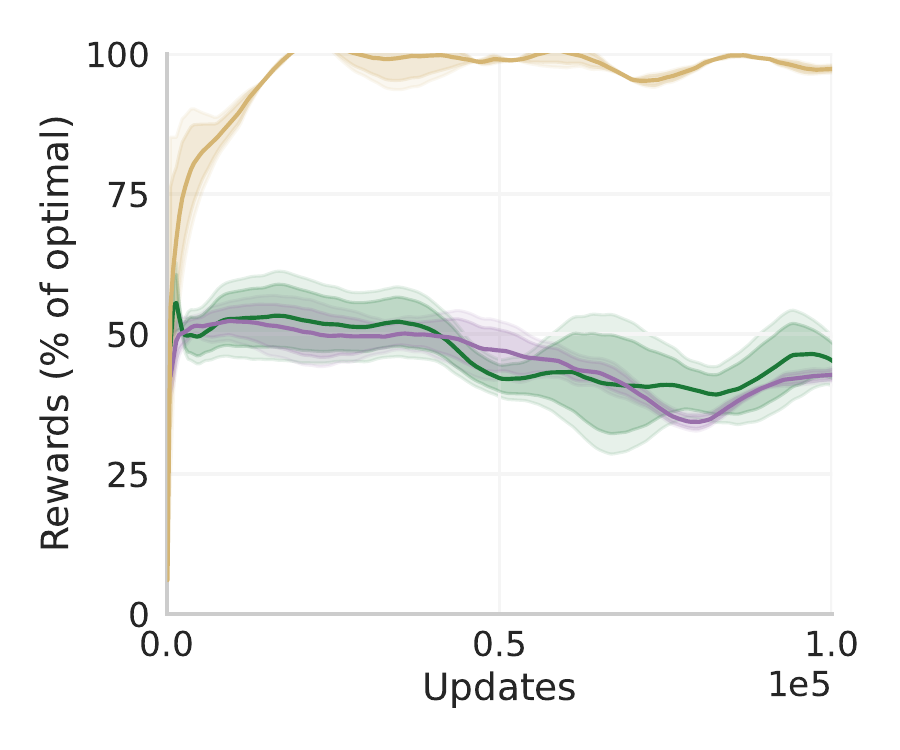}
\caption{195 training DAG structures}\label{app:fig:more_holdouts:even_more}
\end{subfigure}%
\caption{Exploring agent generalization under more restrictive training DAG structure holdouts. In these experiments we impose additional restrictions on the training DAGs, while maintaining the same evaluation set as the original experiments. While the original experiments used 1277 training DAG structures (though of course each structure can be instantiated with infinitely many weight patterns), we impose additional dependency constraints that reduce these to (\subref{app:fig:more_holdouts:more}) 428 or (\subref{app:fig:more_holdouts:even_more}) 195 training DAG structures. In the former case, we still seem some generalization, though it is lower than the original experiments. In the second case generalization performance is quite poor, and we observe evidence of overfitting as training progresses.} \label{app:fig:more_holdouts}
\end{figure}
In this section, we evaluate how agents would perform if they were trained with fewer DAG structures. In the original experiments we held out causal structures that would make node D an ancestor of E. In combination with the other constraints we impose on the graph structures this results in a total of 1277 distinct DAG structures that can occur in training (although of course each structure can in turn be instantiated with many different combinations of edge weights). Here, we reduce this number of possible structures further, by imposing additional constraints:
\begin{itemize}
    \item That neither A nor D can be an ancestor of C or E. This results in 428 possible training DAG structures.
    \item That none of A, D, or E can be an ancestor of B, C, or E. This results in 195 possible training DAG structures.
\end{itemize}
The results of these experiments are shown in Fig. \ref{app:fig:more_holdouts}. As we impose more restrictions on the training distribution, we observe correspondingly weaker generalization. In the first case (428 training structures), we still observe some generalization to novel structures, but in the second case (195 training structures), we observe substantial overfitting.

\subsection{Generalizing to functionally different causal DAGs} \label{app:supp_exp:dag_function_gen}

\begin{figure}[H]
\centering
\begin{subfigure}[b]{0.5\textwidth}
\centering
\includegraphics[width=0.66\textwidth]{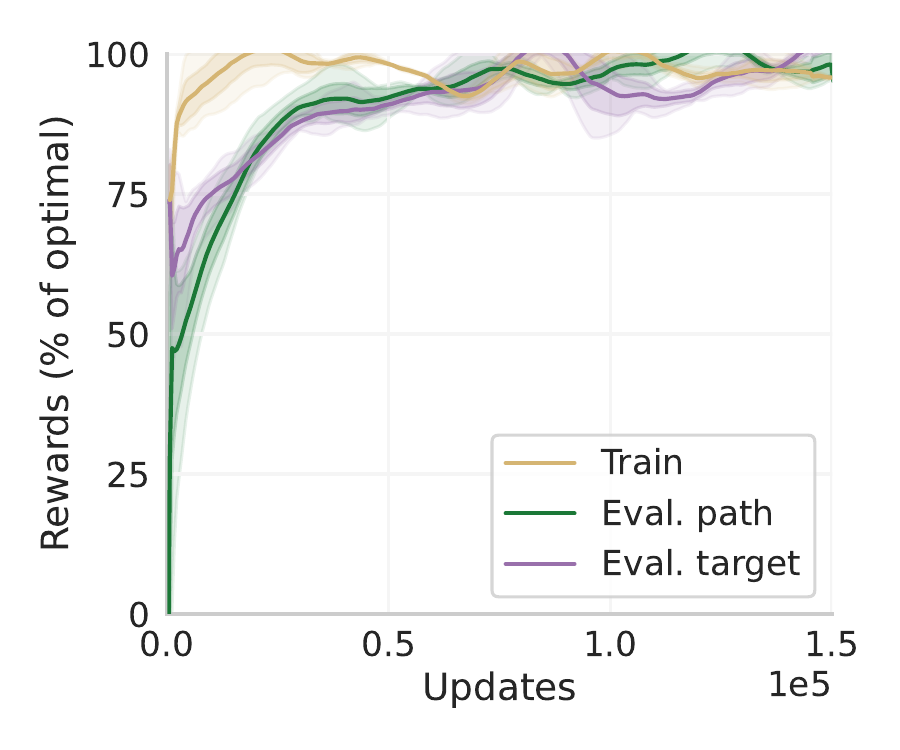}
\caption{Generalizing to nonlinear edges.}\label{app:fig:graph_type_gen:edges}
\end{subfigure}%
\begin{subfigure}[b]{0.5\textwidth}
\centering
\includegraphics[width=0.66\textwidth]{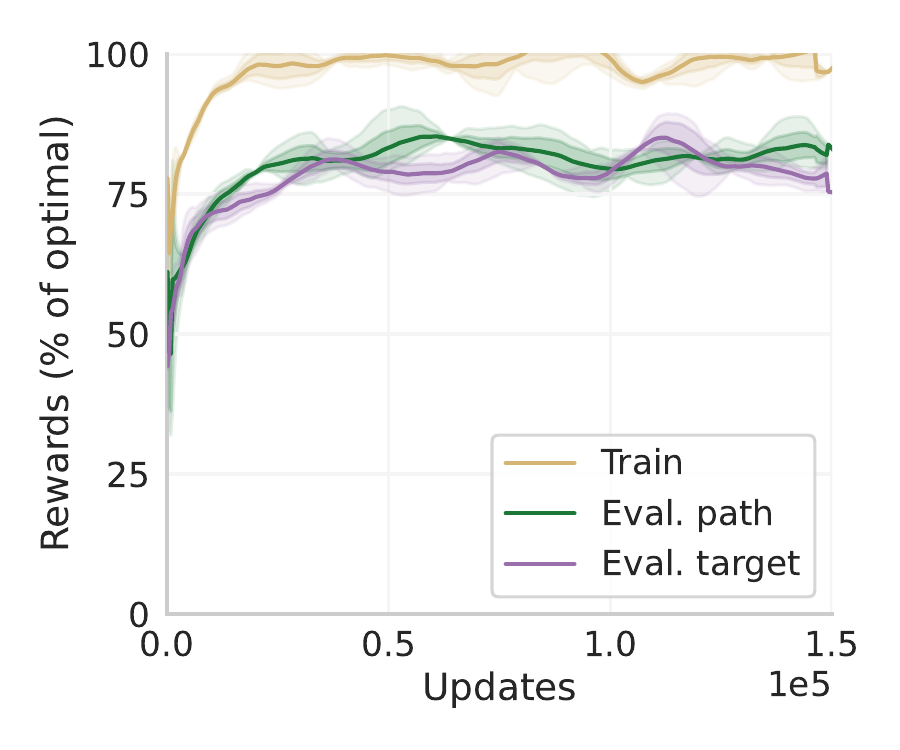}
\caption{Generalizing linear to nonlinear.}\label{app:fig:graph_type_gen:nonlinear}
\end{subfigure}%
\caption{Agents can generalize their causal strategies to different functions types within the causal DAG. In addition to the training/evaluation split used in the main experiments, we introduce an additional shift in evaluation by changing the functions within the graph. (\subref{app:fig:graph_type_gen:edges}) Introducing nonlinearities on the edges in evaluation, in addition to the nonlinearities on the nodes. This has relatively little effect on test performance (perhaps because the node nonlinearities effectively mask most the effect of the edge ones). (\subref{app:fig:graph_type_gen:nonlinear}) Training on linear causal DAGs, and evaluating on nonlinear ones. This evaluation induces a more dramatic shift, and results in a more noticeable generalization gap. However, agents still demonstrate some robustness, achieving over 75\% of optimal performance.} \label{app:fig:graph_type_gen}
\end{figure}

In this experiment, we explored whether agents strategies would be robust to a shift between the kind of causal DAGS experienced in training and evaluation. We considered two shifts: 
\begin{enumerate}
    \item Training on causal DAGs with nonlinearities after the sum on the node (as in the main experiments), then evaluating on DAGs that additionally introduce (leaky ReLU) nonlinearities on the edges before the sum.
    \item Training on linear DAGs (as above in Appendix \ref{app:supp_exp:linear_dag}), and then evaluating on nonlinear DAGs (as in the main experiments).
\end{enumerate}
Fig. \ref{app:fig:graph_type_gen} shows the results. Agent strategies reveal some robustness to these shifts. The agents show very strong generalization in the case of introducing nonlinear edges (though this may be because the nonlinearities on the node mask the effect of edge nonlinearities). In the case of generalizing from linear to nonlinear graphs, there is a larger shift, and so generalization is less perfect, but agents still achieve over 75\% of optimal performance.

\subsection{Adaptive experimentation --- further results} \label{app:supp_exp:adaptive}

In this section, we show further analysis of the agent performance from the adaptive experimentation strategies (Section \ref{sec:experiments:dag:adaptive}) --- specifically, we show that the agent achieves high evaluation rewards, and continues to match the optimal policy much more closely than the baselines. 

It might seem surprising that in this setting the correlation baselines perform very poorly. However this is due to the statistics of the particular situation. Because the number of fully independent variables --- the 5 irrelevant ones --- is equal to the number of interventions performed (and interventions on any other variable do not, of course, affect the irrelevant variables, and often do not affect the target either), there is a relatively high probability that at least one of the independent variables will have a non-trivial correlation with the target. Furthermore, because the variance of the irrelevant nodes is lower on average (as they have no inputs), their correlation is effectively \emph{amplified} in the regression slope, so they end up generally having larger slopes than the actually relevant variables.

\begin{figure}[H]
\centering
\begin{subfigure}[t]{0.305\textwidth}
\includegraphics[width=\textwidth]{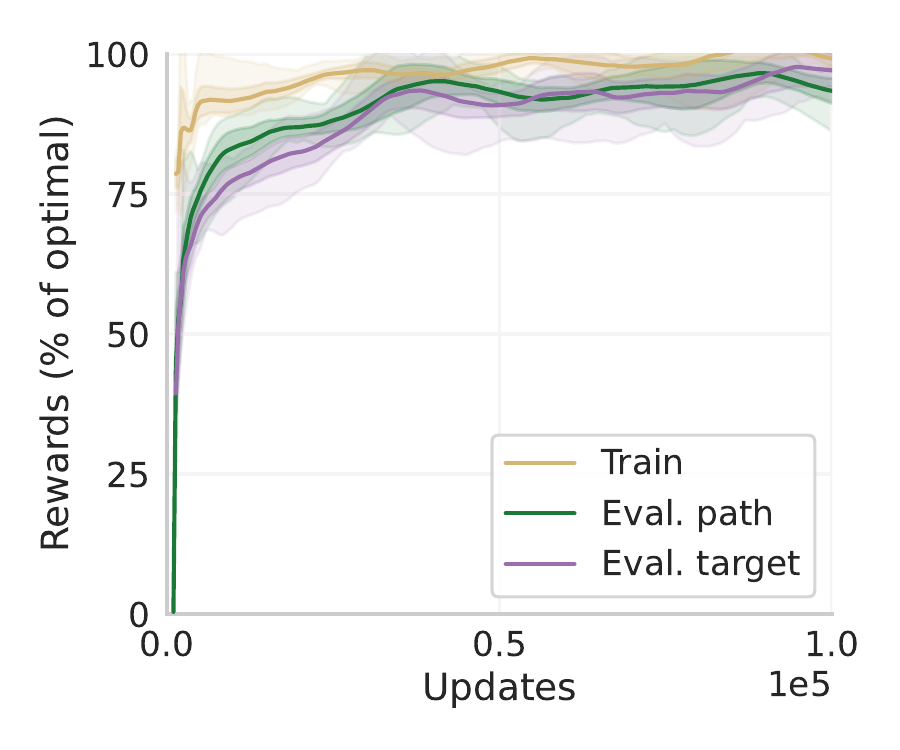}
\caption{Rewards} \label{app:fig:adaptive_dag:rewards}
\end{subfigure}%
\begin{subfigure}[t]{0.445\textwidth}
\includegraphics[width=\textwidth]{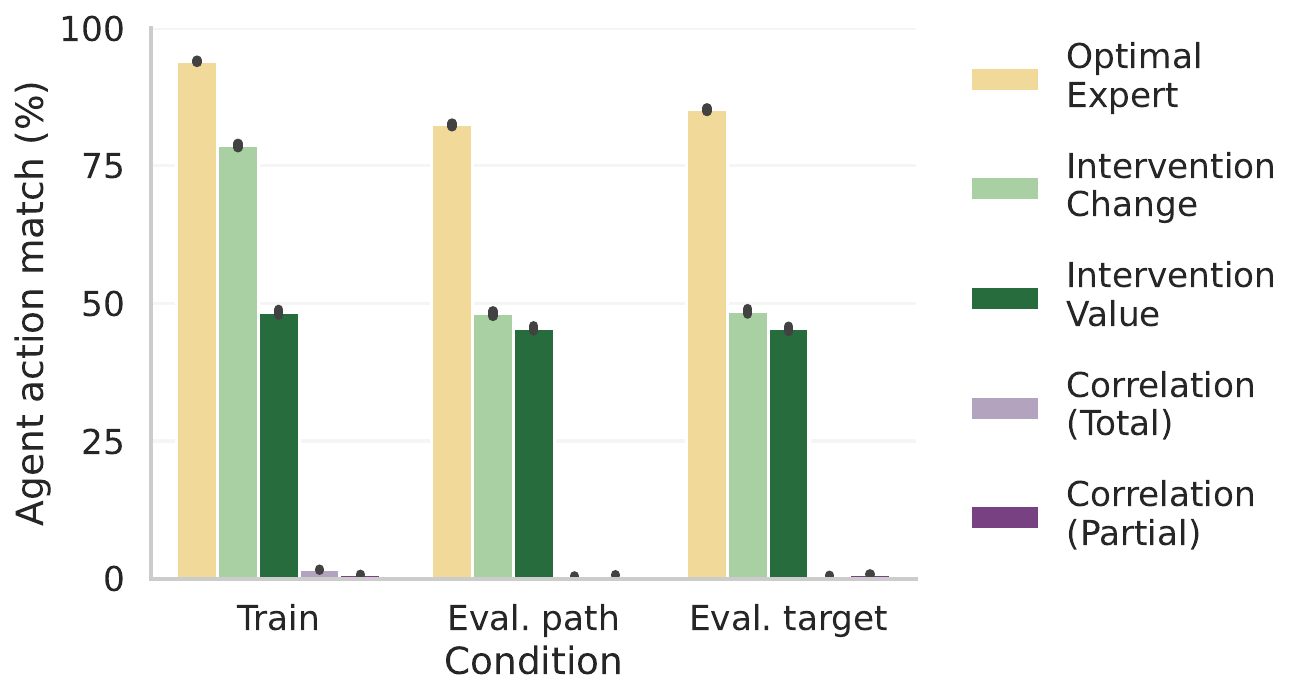}
\caption{Action analysis} \label{app:fig:adaptive_dag:analysis}
\end{subfigure}%
\caption{In the causal DAG environment with adaptive experimentation strategies, (\subref{app:fig:adaptive_dag:rewards}) the agent continues to achieve high rewards in evaluation, and (\subref{app:fig:adaptive_dag:analysis}) the agent continues to more closely align with the optimal policy than with heuristic baselines.}
\label{app:fig:adaptive_dag}
\end{figure}

\subsection{Explanations are key for learning \& generalization in the odd-one-out environments} \label{app:supp_exp:explanations}

\begin{figure}[H]
\centering
\begin{subfigure}[b]{0.4\textwidth}
\centering
\includegraphics[width=\textwidth]{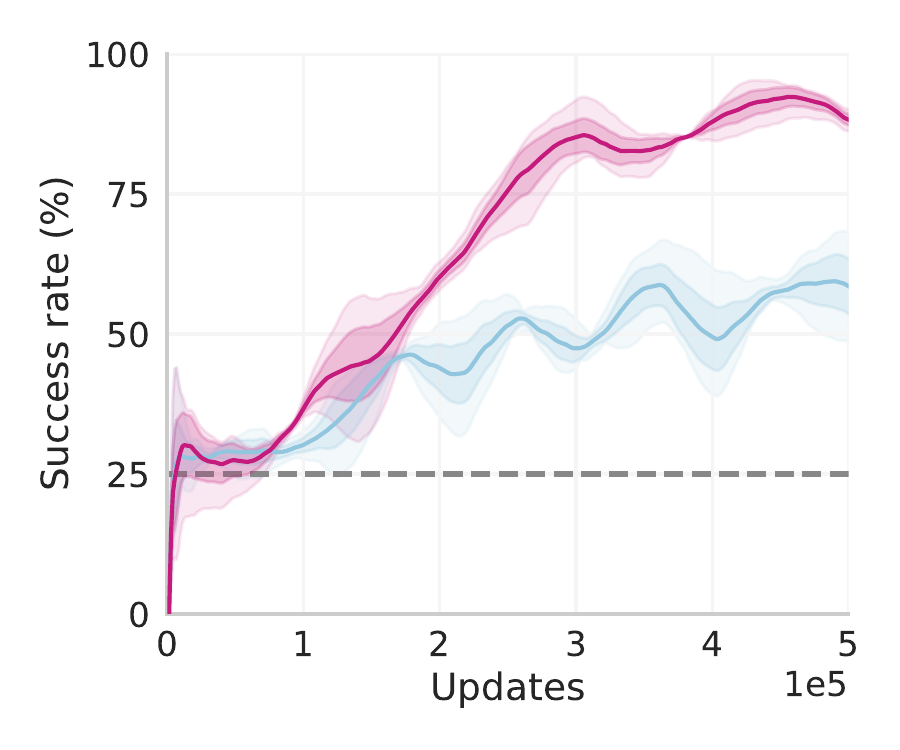}
\caption{Feature set generalization train.}\label{app:fig:explanations:agent:feature:train}
\end{subfigure}
\begin{subfigure}[b]{0.4\textwidth}
\centering
\includegraphics[width=\textwidth]{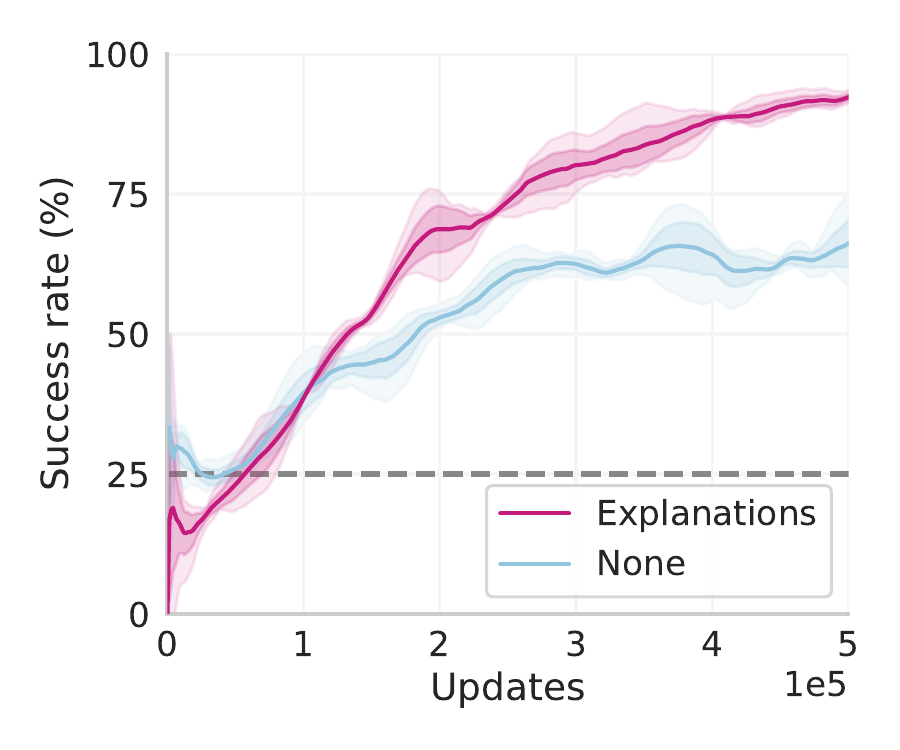}
\caption{Feature set generalization eval.}\label{app:fig:explanations:agent:feature:eval}
\end{subfigure}\\
\begin{subfigure}[b]{0.4\textwidth}
\centering
\includegraphics[width=\textwidth]{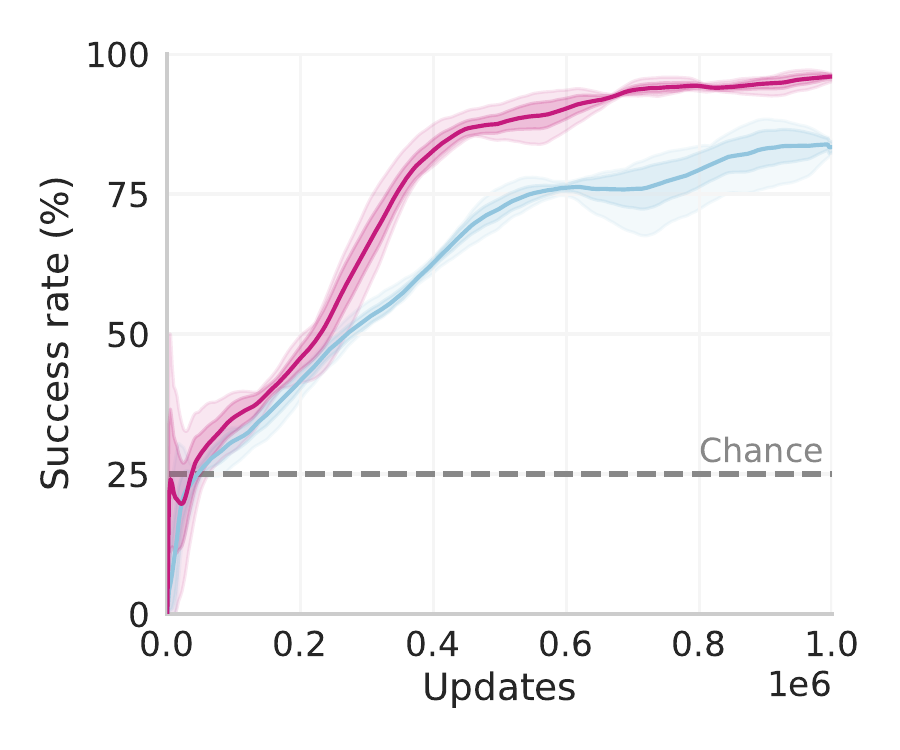}
\caption{Hard dimension generalization train.}\label{app:fig:explanations:agent:dimension:train}
\end{subfigure}
\begin{subfigure}[b]{0.4\textwidth}
\centering
\includegraphics[width=\textwidth]{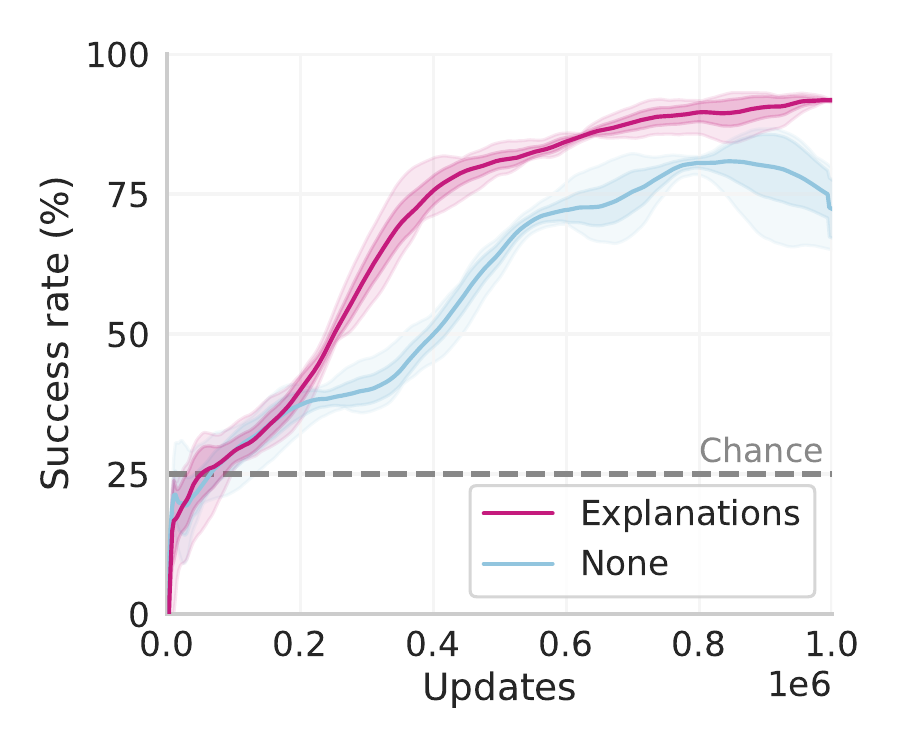}
\caption{Hard dimension generalization eval.}\label{app:fig:explanations:agent:dimension:eval}
\end{subfigure}
\caption{Explanations are key for the basic agent to learn effectively in the odd-one-out environments. Across the two generalization splits: feature splits (top) and hard dimension generalization (bottom), and across train (left) and eval (right) in each case, the agent performs significantly better when trained with explanations.} 
\label{app:fig:explanations:agent}
\end{figure}

In Fig. \ref{app:fig:explanations:agent} we show that agents learn much more effectively with explanations in the odd-one-out environments. Analogously, in the language model experiments explanations were key for generalizing these task structures, as we explore in more detail in the next section.

\subsection{Exploring how prompts affect LM generalization} \label{app:supp_exp:lm_all_explanations}

\begin{figure}[H]
\centering
\includegraphics[width=0.5\textwidth]{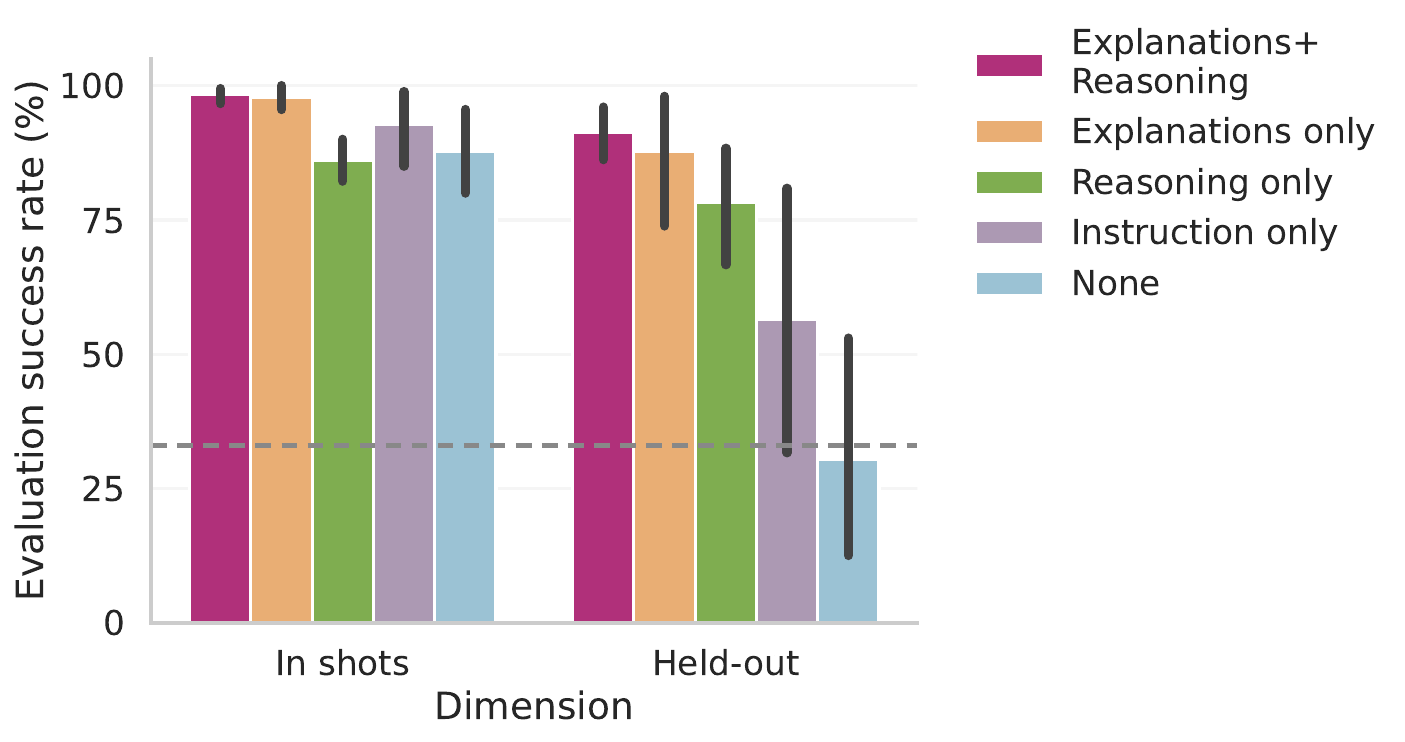}
\caption{Aggregate LM performance across different prompt conditions. Either explanations or chain-of-thought reasoning are sufficient to substantially improve LM average performance on the odd-one-out tasks; however, having both results in more reliable generalization. Instructions may also result in some generalization.} 
\label{app:fig:lm_all_explanations}
\end{figure}

Here, we evaluate language model performance in more detail, across a range of prompt conditions. In addition to the versions with explanations and chain-of-thought reasoning and without either, which we included in the main text, we consider prompts with explanations only, reasoning traces only, or a detailed explanatory instruction included at the beginning of each episode: 
\begin{lstlisting}
    In this game, I must choose an object that is unique along the "correct" dimension. Before choosing an option, I can sometimes transform one of the objects to make it unique along some dimension, not necessarily the correct one:
\end{lstlisting}

In Fig. \ref{app:fig:lm_all_explanations} we show that the LMs can generalize in the odd one out task given either explanations or chain-of-thought reasoning alone; however, performance may be more reliable with both.

\begin{figure}[H]
\centering
\includegraphics[width=\textwidth]{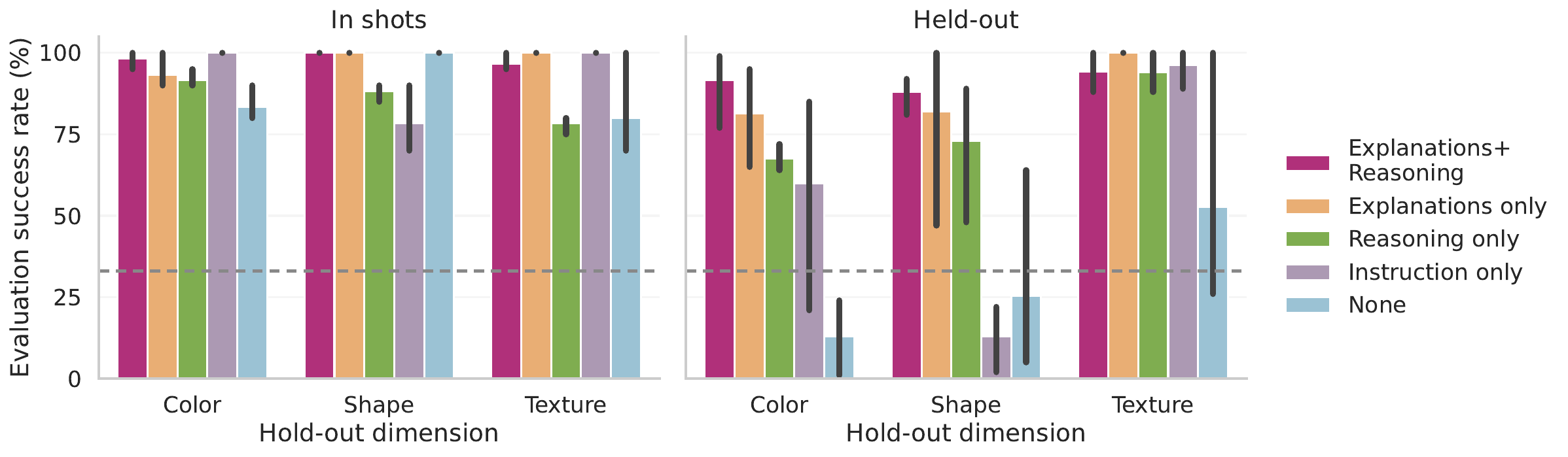}
\caption{LM performance across different prompt conditions, broken down by dimension. There are noticeable differences in generalization across the different dimensions, but the overall pattern of advantages for explanations + reasoning generally holds.} 
\label{app:fig:lm_all_explanations_by_dimension}
\end{figure}

In Fig. \ref{app:fig:lm_all_explanations_by_dimension} we further break these results down by dimension that is held out. Intriguingly, there are striking dfferences in generalization performance on the different dimensions; while prompts with explanations and/or reasoning tend to yield strong generalization across all three, color and shape appear more difficult than texture. These feature biases may be related to those observed in prior work \citep{chan2022transformers}. 

\subsection{Language models generalize less well from a more complex experimentation policy} \label{app:supp_exp:lm_varied_expert}

In the main text, the expert policy for the LM experiments followed a fixed order (color, shape, texture). Here, we explore LM performance if the expert policy in the examples is more complex: it follows the fixed order until it discovers the correct dimension, but then switches to exploiting this knowledge, rather than continuing with its fixed experiments.

\begin{figure}[H]
\centering
\includegraphics[width=0.5\textwidth]{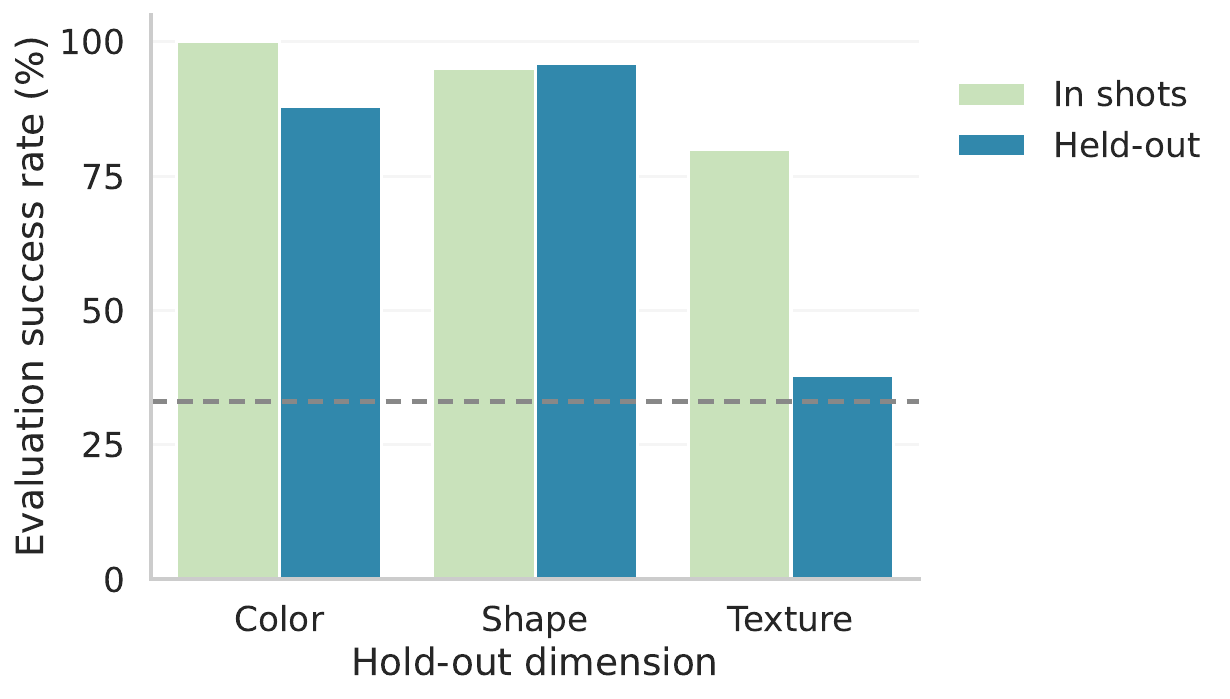}
\caption{Language models generalize less well from an more complex expert policy --- if the expert policy switches to exploiting after it discovers the correct dimension, the language model does not learn or generalize as well in the texture case. (Compare to the results where the expert follows a fixed policy, shown in Fig. \ref{fig:lm}.)} 
\label{app:fig:lm_varied_expert}
\end{figure}

We show the results in Fig. \ref{app:fig:lm_varied_expert}. Performance is noticeably worse, particularly if texture is held out. Texture is the final dimension in the fixed order, so if texture is held out, none of the example shots will contain more than one failed experiment, or any examples of trying texture at all. In this case, the language model performs somewhat less well at new samples even from the dimensions included in the shots (color and shape), and performs drastically worse in held-out evaluation. These results suggest that it is beneficial for the language model to see some shots with multiple experiment failures --- even to perform well on new samples from the color \& shape train set, where multiple failures will not occur if it follows the same exploration order --- and certainly to generalize to the case where texture is held out. It is possible, of course, that more examples in the prompt could overcome this challenge.

\subsection{LMs can generalize the odd-one-out tasks even if an extra object is added to the test trial} \label{app:supp_exp:lm_extra_object}

Evaluating the LM with a surprise extra test object. That is, we kept the prompts the same as the above experiments (three objects in the experimentation and test trials). However, in addition to evaluating on a held-out dimension, we introduced a fourth object in the final test trial of the evaluation episode. More specifically, we kept three objects in the experimentation trials of the test episode, and only surprised the model with a fourth object (which was not unique along any dimension) in the final choice of objects. This experiment tests whether the LM can generalize based on uniqueness when both the relevant dimension and the object distribution changes from training to test.

\begin{figure}[H]
\centering
\includegraphics[width=0.5\textwidth]{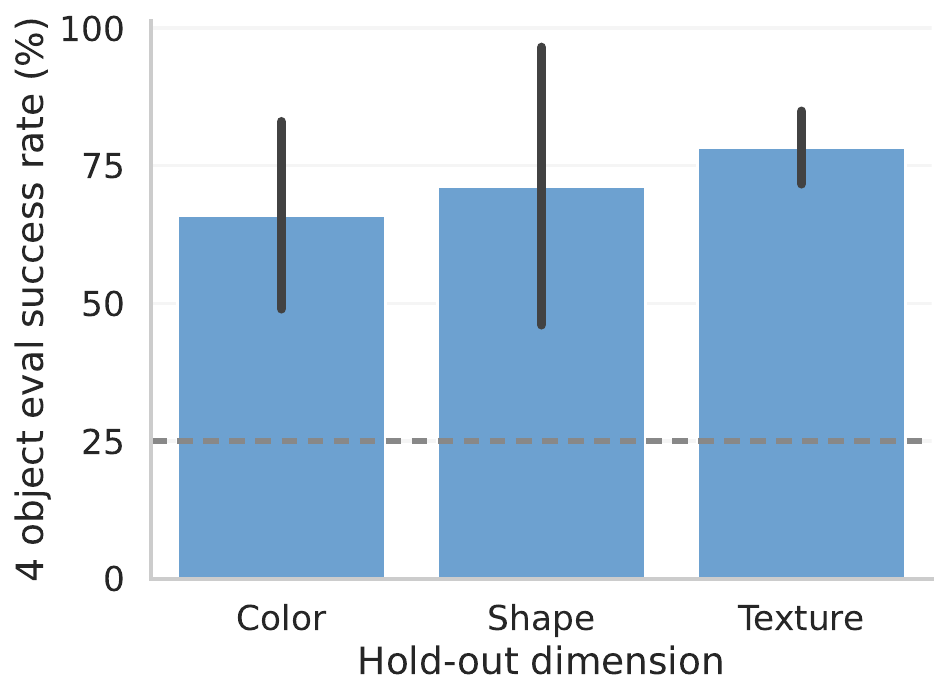}
\caption{LM evaluation performance on held-out dimensions, as above, but with an extra surprise object in the final test trial. The LM performs far above chance even in this more challenging test condition.} 
\label{app:fig:four_object_test}
\end{figure}

The results are shown in Fig. \ref{app:fig:four_object_test}. While the added object does reduce generalization performance somewhat relative to the original three-object version, performance is still far above chance (25\%). Thus, the LMs appear to generalize relatively robustly when prompted with explanations, even if the test trial differs from the training in several ways.

\clearpage
\section{Supplemental methods}

\subsection{Agent hyperparameters \& training}

In Table \ref{tab:app:hyper} we list the architectural and hyperparameters used for the main experiments. All agents were implemented using JAX \citep{jax2018github} and Haiku \citep{haiku2020github}.

\begin{table}[H]
    \centering
    \caption{Hyperparameters used for the agents.}
    \vspace{0.4em}
    \begin{tabular}{|c|c|c|}
      \hline
      & \textbf{Causal DAG} & \textbf{Odd-one-out} \\ \hline
      \hline
      Memory feature dimension & \multicolumn{2}{c|}{512} \\ \hline
      Memory layers & \multicolumn{2}{c|}{4} \\ \hline
      Memory num. heads & \multicolumn{2}{c|}{8} \\ \hline
      TrXL extra length & 16 & 128 \\ \hline
      \hline
      Visual encoder &  & CNN\\ \hline 
      Vis. enc. channels &  & (16, 32, 32) \\ \hline 
      Vis. enc. filt. size & & (9, 3, 3)\\ \hline 
      Vis. enc. filt. stride & & (9, 1, 1)\\ \hline 
      \hline
      Policy & \multicolumn{2}{c|}{Single linear layer.} \\ \hline
      Explanation decoder & & 1-layer LSTM, embedding size 128 \\ \hline
      \hline
      BC loss weight & \multicolumn{2}{c|}{1.} \\ \hline
      Explanation loss weight & & 0.1 \\ \hline
      \hline
      Batch size & \multicolumn{2}{c|}{32} \\ \hline
      Training trajectory length & 20 & 50 \\ \hline
      \hline 
      Optimizer & \multicolumn{2}{c|}{Adam \citep{kingma2014adam}} \\ \hline
      LR & \(2\cdot10^{-4}\) & \(1\cdot10^{-4}\) \\ \hline
      Max gradient norm & 10 & 1 \\ \hline
    \end{tabular}
    \label{tab:app:hyper}
\end{table}

Hyperparameters were not extensively tuned; they were mostly set to the same values from prior work \citep{lampinen2022tell}, with the following exceptions:
\begin{itemize}
    \item We increased the weight on the explanations loss in the intervention odd-one-out environments (to 0.1), to better match the magnitude of the BC loss; learning was slightly impaired with a smaller value of the explanation loss.
    \item Learning in the intervention odd-one-out environments was unstable in our initial experiments --- the loss would sometimes become NaN --- so we decreased the max value to which the gradient norm was clipped, which seemed to resolve the issue.
    \item We used shorter trajectory length in the simple causal DAG environment.
    \item We used a simpler policy architecture (single linear layer).
\end{itemize}

\subsection{Compute}

All agents were trained using a \(2 \times 2\) TPU v2 or v3 slice. Language model inference was done on \(2 \times 4\) TPU v4 slices. Total usage across all experiments (both agents and LMs), including preliminary runs etc., was approximately 450 TPU-hours.

\end{document}